\PassOptionsToPackage{nameinlink,capitalise,noabbrev}{cleveref} 
\documentclass[9pt,a4paper,twocolumn,twoside]{gc/article}
\usepackage{microtype}
\usepackage{graphicx}
\usepackage{subcaption}
\usepackage{booktabs}

\usepackage{amsmath}
\usepackage{amssymb}
\usepackage{array}
\usepackage{listings}
\usepackage{inconsolata}  
\usepackage{mathtools}
\usepackage{upgreek}

\newcommand{\ourformat}{\texttt{S3D8}}
\newcommand{\errorrange}[2]{%
\shortstack[c]{\\[1pt]#1\\[0pt]{\color{gray}\scriptsize #2}}%
}
\newcolumntype{C}[1]{>{\centering\arraybackslash}m{#1}}

\crefname{lstlisting}{listing}{listings}
\Crefname{lstlisting}{Listing}{Listings}

\lstdefinestyle{promptstyle}{
  basicstyle=\ttfamily\footnotesize,
  breaklines=true,
  breakautoindent=false,
  breakindent=0pt,
  breakatwhitespace=false,
  columns=fullflexible,
  keepspaces=true,
  numbers=none
}

\journalname{}
\title{Llama-Mobile: Efficient 2.7-Bit Quantization of VLMs}
\correspondingauthor{Douglas Orr}{1}{douglaso@graphcore.ai}
\author{Luka Ribar$^{\,1}$, Jeevan Bhoot$^{\,2}$ and \thecorrespondingauthor}
\institution{{$^1$}Graphcore Research \quad {$^2$}Arm}

\begin{abstract}
Deploying vision-language models (VLMs) on mobile devices is challenging due to their significant memory and compute requirements. We present a framework for quantizing VLMs for efficient inference on resource-constrained hardware. Our approach combines a quantization pipeline that uses the model itself to generate training data and does not require access to the training setup, with a novel $2.7$-bit-per-parameter format supporting efficient execution on Arm CPUs. We validate our approach by compressing the Llama 3.2 11B Vision Instruct model to $3.7$ GB with $8$-bit activations, preserving strong performance on a set of standard visual question answering tasks.
\end{abstract}

\begin{document}

\maketitle
\thispagestyle{firststyle}
\gcabstract

\section{Introduction}

Large language models (LLMs) have recently demonstrated strong performance across difficult language tasks requiring complex reasoning and reading comprehension \cite{llama3, gemini2.5, qwen3, deepseek3.2}. Building on these capabilities, vision-language models (VLMs), such as Llama 3.2 Vision, Gemma 3, and Qwen3-VL \cite{llama3, gemma3, qwen3vl}, add a vision encoder to the language model backbone, enabling the LLM to process and understand both image and text inputs. Such models are a natural fit for mobile applications, but current systems often remain too computationally and memory-heavy for mobile deployment.

Quantization is a practical way to reduce model memory usage and improve inference efficiency with minimal performance degradation. However, aggressive low-bit quantization of VLMs remains difficult in practice. In addition to selecting a numerical format that minimizes model degradation \cite{dettmers2023scaling,orr2025optimal}, the quantization procedure itself is important \cite{frantar2022gptq, liu2024llmqat}. Quantization-aware training (QAT) is highly effective, but downstream performance is sensitive to training data selection \cite{liu2024llmqat, du2025optimize_quantization}.

In this work, we address two practical constraints for VLM quantization: achieving sub-$3$-bit weight compression with efficient Arm CPU execution, and doing so without access to the original training data. First, we introduce \ourformat{}, a novel $2.7$-bit-per-parameter quantization format that packs three signed weights into each byte through a shared centroid index and decodes them to \texttt{INT8} for inference. Second, because such aggressive compression can substantially degrade model quality, we use a QAT procedure based on the model's own generations. As previous similar approaches are not directly applicable to the multimodal setting \cite{liu2024llmqat}, we introduce a simple data-generation pipeline that preserves question-answering performance without directly fine-tuning on downstream tasks.

\begin{figure}[t]
  \centering
  \includegraphics[width=\columnwidth]{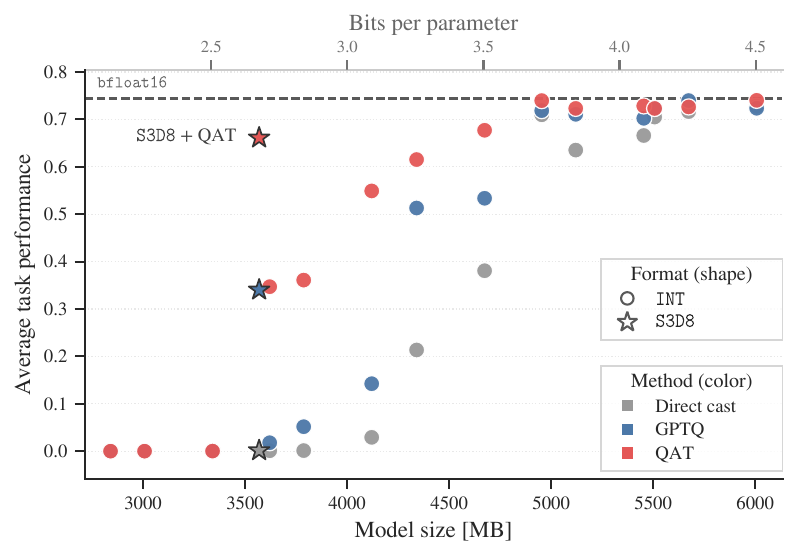}
  \caption{Trade-off between overall task performance and model compression under direct casting, GPTQ, and QAT. Under our QAT scheme, \ourformat{} is approximately $22\%$ smaller than block-scaled \texttt{INT} at matched task performance in our evaluation. Directly casting model weights to fewer than $3.5$ bits per parameter causes significant model degradation. While both GPTQ and QAT can improve performance, QAT is superior at fewer bits per parameter. The dashed line shows the original \texttt{bfloat16} ($21.3$ GB) model performance.}
  \label{fig:overview}
\end{figure}

We apply our framework to the Llama 3.2 11B Vision Instruct model \cite{llama3} by compressing its weights to $3.7$ GB while quantizing activations to $8$-bit, showcasing limited degradation on a set of standard visual-question answering tasks, as summarized in \cref{fig:overview}. Our evaluation compares QAT with direct casting and GPTQ \cite{frantar2022gptq}, a strong PTQ baseline, to assess the effects of both the numerical format and the quantization procedure.

\vspace{1em}
\paragraph{Contributions}
\begin{itemize}
\item We present a simple and generalizable QAT pipeline for multimodal models that does not require access to the original training data and enables effective sub-$3$-bit quantization with limited model degradation.
\item We introduce a novel $2.7$-bit-per-parameter quantization format that enables efficient Arm CPU inference under aggressive model compression.
\item We provide a full VLM inference implementation for Arm CPUs on Linux and Android.
\end{itemize}

We provide the full quantization pipeline and inference implementation at: \href{https://github.com/graphcore-research/llama-mobile-vlm}{graphcore-research/llama-mobile-vlm}.

\section{Related Work}

\paragraph{Quantization}

Quantized weights can be optimized either through post-training quantization (PTQ) techniques \cite{frantar2022gptq, xiao2023smoothquant} or QAT \cite{liu2024llmqat}. While PTQ offers a less resource-intensive alternative, QAT is generally more effective, especially at low-bit widths, where quantization error tends to be significant \cite{liu2025paretoq}. The quantized values may be represented with uniform integer grids, floating-point values, or lookup tables, often accompanied with block, channel, or tensor-scaling, as well as vector quantization codes, including sign-factorized codebooks \cite{dettmers2023scaling, dettmers2023qlora, rouhani2023microscaling, egiazarian2024aqlm, son2025nsnquant}. Recent VLM-oriented quantization work focused on effectively applying PTQ techniques in the multimodal setting \cite{wang2024qvlm, li2025mbq, xue2025vlmq}. Our work instead studies QAT for aggressive sub-$3$-bit VLM weight compression and treats the numerical format as part of the hardware co-design.

\paragraph{Format co-design}

Hardware-aware format design and format-aware kernel design are important avenues for practical formats. \textcite{gope2025kernels} explore weight layout, codebooks, and SIMD kernels for Arm CPUs, with practical implementation in \texttt{llama.cpp} \cite{llamacpp}. Our work employs similar techniques, but \ourformat{} uses a vector format with potential for a better rate--distortion trade-off than non-uniform scalar formats. In the extreme low-bit regime, Sherry \cite{huang2026sherry} uses a SIMD-friendly format for ternary quantization with structured sparsity at $1.25$ bits per parameter. More broadly, hardware co-design has focused on block formats such as MXFP and NVFP4 \cite{ocpspec,xin2026qad-nvfp4}, which reduce the amount of computation in high precision. Similarly, DeepSeek V3 adapts their block scaling format to the available hardware by using large $2$D blocks of weights to match GEMM hardware tile size \cite{deepseek2024}. Our work differs in that we co-design the numerical format around the SIMD decode path on Arm CPUs.

\section{Method}
\subsection{Overview}

Our goal is to compress a pretrained vision-language model into a low-bit representation that can be effectively used for inference on low-resource Arm CPUs, while preserving the model's visual question-answering ability. We focus on a practical setting where we do not have access to the model's original training data or recipe. This is a particularly challenging problem, as aggressive quantization can lead to severe model degradation, especially in a multimodal setting.

To achieve an effective high level of compression, we use quantization-aware training with knowledge distillation, where the original full-precision model serves as a teacher and the quantized model is trained to match its outputs. Since the effectiveness of QAT depends strongly on the training data distribution, we construct a synthetic multimodal training set using the model itself.

Once the target quantization format is specified, we apply a consistent QAT procedure, enabling direct comparison between different numerical formats in our ablations. To enable efficient inference on Arm CPUs, we introduce a $2.7$-bit-per-parameter weight format that provides a good balance between compression, model quality, and execution speed.

The overall pipeline thus has the following components:
\begin{enumerate}
    \item \textbf{Synthetic Data Generation}: Using the model itself, generate the training samples to be used for quantization-aware training.
    \item \textbf{Quantization-Aware Training}: After choosing the numerical format of the quantized model, distill the original model using the synthetic data.
    \item \textbf{Mobile Deployment}: Execute the model on Arm CPU, using the trained weights from the QAT stage.
\end{enumerate}

The remainder of this section first defines the common QAT objective used throughout the paper (\cref{sec:qat}), then describes the two components central to our approach: the synthetic data-generation pipeline (\cref{sec:qat_data_pipeline}) and the proposed low-bit numerical format (\cref{sec:format}).

\subsection{Quantization-Aware Training}
\label{sec:qat}

We now define the QAT setup used throughout this work. Our goal is to train a quantized \textit{student} model to match the behavior of the original full-precision \textit{teacher} model on a synthetic multimodal dataset.

Let
\begin{equation*}
\mathcal{D} = \{(I_i, x_i, y_i)\}_{i=1}^{M}
\end{equation*}
be the training dataset of size $M$, where $I_i$ denotes an image input, $x_i$ a text prompt, and $y_i = (y_{i,1}, \dots, y_{i,|y_i|})$ a teacher-generated response sequence.

We define a teacher model $f_T(\cdot; \theta_T)$, parameterized by $\theta_T \in \mathbb{R}^{P}$, operating in high precision (\texttt{bfloat16}), and a student model $f_S(\cdot; \theta_S)$, parameterized by $\theta_S \in \mathbb{R}^{P}$, trained using QAT. Let $Q(\cdot)$ denote the quantization operator corresponding to the target numerical format. During training, the student forward pass uses quantized parameters
\begin{equation*}
\hat{\theta}_S = Q(\theta_S),
\end{equation*}
while gradients are propagated through $Q$ using a straight-through estimator \cite{bengio2013ste}.

For a sample $(I, x, y)$, both teacher and student define a next-token distribution at each decoding step $t$, conditioned on the image, the prompt, and the previously generated response tokens $y_{<t}$:
\begin{equation*}
p_T^{(t)} = f_T(I, x, y_{<t}; \theta_T), \qquad
p_S^{(t)} = f_S(I, x, y_{<t}; \hat{\theta}_S).
\end{equation*}
Here, $p_T^{(t)}$ and $p_S^{(t)}$ are distributions over the token vocabulary for the $t$-th response token.

We train the student to match the teacher distribution on the generated response tokens only. We define the per-sample distillation loss as
\begin{equation*}
\mathcal{L}_{\mathrm{KD}}(I,x,y)
=
\frac{1}{|y|}
\sum_{t=1}^{|y|}
D_{\mathrm{KL}}
\left(
p_T^{(t)}
\,\|\,
p_S^{(t)}
\right),
\end{equation*}
where $|y|$ is the number of generated response tokens in the sample, and $D_{\mathrm{KL}}$ is the Kullback–Leibler divergence.

Minimizing this objective encourages the quantized student to preserve the teacher's multimodal generation behavior under the target low-bit representation. The remaining challenge is therefore to construct a synthetic training set $\mathcal{D}$ that captures the relevant image-conditioned question-answering behavior of the original model.

\subsection{Synthetic QAT Data Pipeline}
\label{sec:qat_data_pipeline}

The main challenge in constructing data for multimodal quantization-aware training is that there is no direct analogue of the large-scale generic text corpora commonly used for language-model pretraining. While many multimodal datasets contain large numbers of image-caption pairs, such data is not necessarily well-matched to the behavior that an instruction-tuned VLM is expected to preserve. In particular, short descriptive captions provide only limited supervision for long-form generation, question answering, and instruction following, all of which are important for downstream VLM performance.

Ideally, the QAT dataset should consist of images paired with diverse textual responses of varying length and style. The data should expose the model to a range of different response types, including direct visual descriptions, short factual answers, as well as longer explanatory responses, while remaining sufficiently generic to avoid overfitting to particular task types. Although multimodal instruction datasets exist \cite{liu2023llava,dai2023instructblip}, they may not match the original model's response style or prompt format. We therefore use teacher generations as the text targets for distillation.

For the image examples, we use ImageNet \cite{imagenet}, which provides a large and diverse collection of natural images, while remaining generic and independent of the downstream evaluation tasks. We then query the teacher model on these images, saving the model's generated responses as the target sequences for knowledge distillation. Each final QAT sample consists of an image, a sampled prompt, and the corresponding teacher-generated response sequence.

\paragraph{Prompt Sampling} The prompting strategy here is essential; a fixed set of prompts can lead to a limited variety in the model's outputs, hindering downstream generalization (see \cref{fig:prompt_overall}). To address this, our procedure randomly samples prompts in the following way:

\begin{itemize}
    \item We randomly apply the instruction-tuned prompt template with high probability $p_{\mathrm{temp}}$; this ensures most of the examples reinforce the fine-tuned model behavior, while also retaining the pretrained behavior that does not use the
    instruction template.
    \item The prompt pool contains general image-comprehension questions that can be applied across most natural images.
    \item With probability $p_{\mathrm{inst}}$, we prepend an instruction block containing a sampled response guideline, with optional constraints on answer length, output format, or instruction adherence.
    \item Length constraints are biased toward short and medium-length responses, with a small fraction of instructions enforcing a longer model response.
\end{itemize}

Importantly, this prompting strategy is generic and does not use the downstream task data or benchmark-specific prompts in order to avoid overfitting to particular question-answering tasks.

\subsection{The \ourformat{} Format}
\label{sec:format}

\begin{figure*}
    \centering
    \begin{minipage}[t]{0.67\textwidth}
    \centering
    \includegraphics[width=\linewidth]{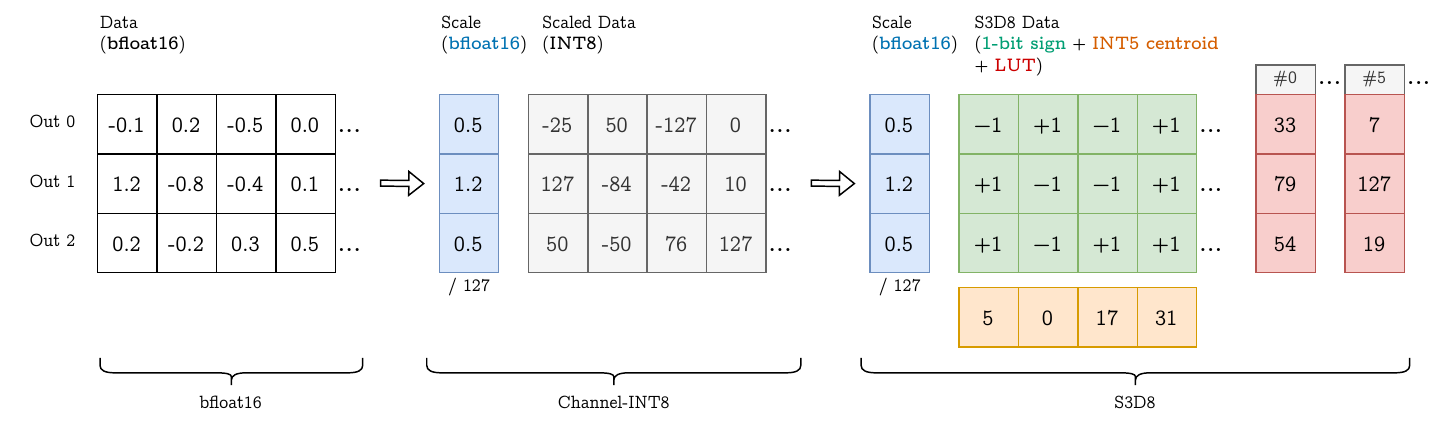}
    \end{minipage}
    \hspace{0.02\textwidth}
    \begin{minipage}[t]{0.30\textwidth}
    \centering
    \includegraphics[width=\linewidth]{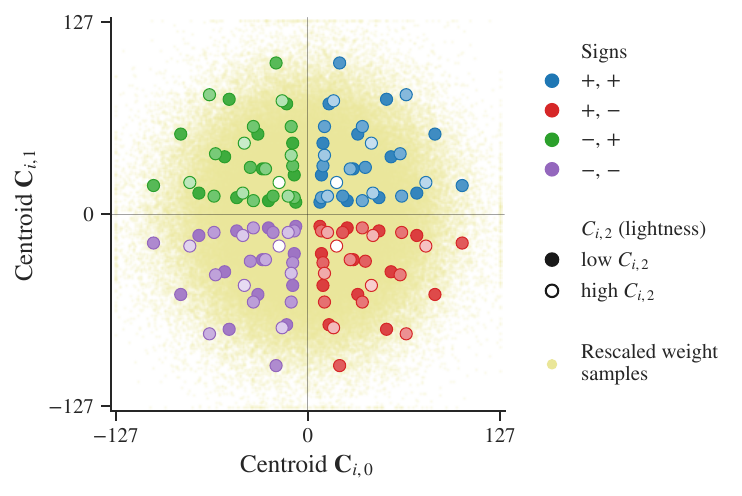}
    \end{minipage}
    \caption{
    \emph{Left:} The \ourformat{} encoding flow. Starting from \texttt{bfloat16} data, a scale is chosen based on the absolute maximum of each output channel. Rounding and quantizing the scaled elements produces the channel-\texttt{INT8} compute format. This is further compressed to \ourformat{} by extracting the sign and storing the index of the closest centroid. Note that this logical view does not portray the physical layout of the data and lookup tables (see \cref{fig:s3d8-bit-layout}).
    \emph{Right:} \ourformat{} centroids trained on unit Normal weights. The $32$ centroids in \textbf{\textcolor[HTML]{1F77B4}{blue}} are trained on the absolute values of the rescaled weights. The $3$ sign indicator bits provide the same centroids reflected into each of the $8$ orthants ($4$ shown).
    }
    \label{fig:s3d8}
\end{figure*}

Practical weight storage formats must support efficient dequantization on the target device. To this end, we introduce the \textit{signed $3$D $8$-bit} (\ourformat{}) format, designed for fast dequantization to \texttt{INT8} on Arm CPUs using logical operations and small lookup tables.

\paragraph{\ourformat{}} The format stores $3$ weights in $8$ bits, using a shared $5$-bit centroid index and $3$ sign indicator bits (one per weight), illustrated in \cref{fig:s3d8}. Conceptually, a single $32$-element table lookup gives the absolute values of a vector of $3$ weights; these are multiplied by their sign bits and per-channel scales to reconstruct them. With channel scaling, the dequantized weight matrix $\widetilde{W} \in \mathbb{R}^{n \times k}$, where each group of $3$ output channels shares a centroid index, is given by
\begin{equation}\label{eqn:s3d8-dequant}
    \widetilde{W}_{ij} = \alpha_i \cdot s_{ij} \cdot C_{ q_{\lfloor i/3\rfloor,j} ,\; i \bmod 3} \,,
\end{equation}
with per-channel scales $\boldsymbol{\alpha} \in \mathbb{R}^{n}$, per-element signs $\boldsymbol{s} \in \{-1, 1\}^{n \times k}$, per-matrix centroids $\mathbf{C} \in \{0,\ldots,127\}^{32 \times 3}$, and quantization indices $\mathbf{q} \in \{0,\ldots,31\}^{\lceil n/3 \rceil \times k}$. We assume $0$-based indexing, i.e. starting at $W_{00}$.

With channel scales stored in \texttt{bfloat16}, the average size for $\widetilde{\mathbf{W}}$ is approximately $(8/3 + 16/k)$ bits/element. For sufficiently large $n$ and $k$, centroid storage and padding of $n$ to a multiple of $3$ are negligible.

\paragraph{Physical layout \& decoding}
Our format employs three design choices to improve decoding efficiency. The first is to pack separate output channels into each byte, rather than consecutive input weights. This reduces the need for complex padding and alignment logic when used in a fused dequantization and dot product kernel. Second, we exploit the $64$-entry table lookup support in Arm CPUs to look up \textit{signed} values directly, combining the shared centroid index with the sign of an individual element to construct the lookup index. Third, we reorder and combine sign indicator bits to require just $5$ logical instructions to construct the $3$ lookup indices. These techniques are illustrated in \cref{fig:s3d8-bit-layout} and described in more detail in \cref{sec:app-format-details}.

\paragraph{Quantization}
We separate quantization into two phases. First, centroids are constructed once before performing QAT. Second, during QAT, channel scales, quantization indices and signs are updated while centroids are kept frozen.

To construct centroids, we first normalize each weight matrix by the absolute maximum of each output channel and then take absolute values. The matrix is then zero-padded and reshaped into a set of three-element vectors. We apply the Lloyd-Max algorithm \cite{lloyd1982kmeans} with k-means++ initialization \cite{arthur2007kmeans++} to obtain centroids that minimize squared reconstruction error:
\begin{align*}
    \mathbf{C} &= \mathrm{Round}\left(\mathrm{Lloyd\text{-}Max}( \{\mathbf{W}_{i^{\prime}j\,\cdot}^{\prime}\}_{i^{\prime}\!,j} ) \right) \\
    \text{where} \quad
    W_{i^{\prime}jd}^{\prime}
        &= \frac{|W_{(3 i^{\prime}+d),j}| }{ \alpha_{3 i^{\prime}+d}}
            \,,\; d \in \{0,1,2\} \,. \\
    \alpha_i &= \max_j | W_{ij} | \,/\, 127
\end{align*}
See \cref{fig:s3d8} for an example on artificial data.

In the QAT phase, we quantize the weight matrix in the forward pass given these centroids. First, the scale is updated as above, and the sign of each weight is recorded. The quantization index is assigned to the closest centroid to each scaled-absolute $3$-vector:
\begin{align*}
    s_{ij} &= \begin{cases}
        +1 & W_{ij} \ge 0 \\
        -1 & W_{ij} < 0
    \end{cases} \\
    \mathbf{q}_{i^{\prime}j}
        &= \arg\min_{l \in \{0,\ldots,31\}} \lVert \mathbf{C}_l - \mathbf{W}_{i^{\prime}j}^{\prime} \rVert_2
        \,.
\end{align*}
The weight is then reconstructed according to \cref{eqn:s3d8-dequant}.

\section{Results}
All experiments use the Llama 3.2 11B Vision Instruct model, whose original parameters are in \texttt{bfloat16}. We evaluate three quantization procedures: direct casting, GPTQ \cite{frantar2022gptq}, and QAT. For each quantized variant, we quantize all language and vision weights except one-dimensional parameters such as biases. The direct-cast and QAT variants use per-channel-scaled \texttt{INT8} activations, whereas GPTQ is weight-only and retains \texttt{bfloat16} activations.

We compare \ourformat{} against three scalar quantization baselines at matched or similar model sizes. \texttt{INT} uses uniformly spaced signed integer levels. \texttt{lloyd-max} uses non-uniform scalar levels obtained by minimizing squared reconstruction error for the weight distribution. \texttt{student-t} uses scalar quantization levels optimized for a Student-$t$ prior, following \textcite{orr2025optimal}. \texttt{lloyd-max} and \ourformat{} use channel-scaling, while \texttt{INT} and \texttt{student-t} use block-scaling with varying block sizes. In all cases the scale is stored in \texttt{bfloat16} and set according to the absolute maximum of the channel or block. All numerical-format baselines use the same QAT procedure as \ourformat{}, allowing the comparison to isolate the effect of the numerical format.

We benchmark the models on fixed $1024$-example subsets from four visual question-answering tasks: VQAv2 \cite{vqav2}, ChartQA \cite{chartqa}, DocVQA \cite{docvqa}, and AI2D \cite{ai2d}, reporting individual metrics and the overall average.

Full details of the training hyperparameters, numerical formats, and evaluation can be found in \cref{app:experimental_details}.

\begin{table*}[t]
\centering
\caption{Comparison of QAT formats at similar model sizes. \ourformat{} achieves favorable downstream performance against similarly-sized \texttt{INT}, \texttt{student-t}, and \texttt{lloyd-max} formats. VQAv2, ChartQA, and AI2D report accuracy, while DocVQA reports ANLS. Note that the model sizes here do not include vocabulary and metadata. Further results can be found in \cref{app:additional_results}.}
\label{tab:format-comparison}
\begin{tabular}{lrrrrrrr}
\toprule
Format & Size [MB] & bpp & VQAv2 & ChartQA & DocVQA & AI2D & Avg. \\
\midrule
\texttt{bfloat16} & $21{,}340$ & $16.00$ & $0.754$ & $0.747$ & $0.844$ & $0.631$ & $0.744$ \\
\midrule
\textbf{\ourformat{}} & $\mathbf{3{,}569}$ & $\mathbf{2.68}$ & $\mathbf{0.702}$ & $\mathbf{0.648}$ & $\mathbf{0.740}$ & $\mathbf{0.554}$ & $\mathbf{0.661}$ \\
\texttt{INT} & $3{,}620$ & $2.71$ & $0.579$ & $0.303$ & $0.258$ & $0.249$ & $0.347$ \\
\texttt{student-t} & $3{,}620$ & $2.71$ & $0.647$ & $0.598$ & $0.670$ & $0.344$ & $0.565$ \\
\texttt{lloyd-max} & $3{,}459$ & $2.59$ & $0.638$ & $0.510$ & $0.444$ & $0.151$ & $0.436$ \\
\bottomrule
\end{tabular}
\end{table*}

\begin{figure*}[t]
  \centering
  \begin{subfigure}[t]{0.49\textwidth}
    \centering
    \includegraphics[width=0.8\linewidth]{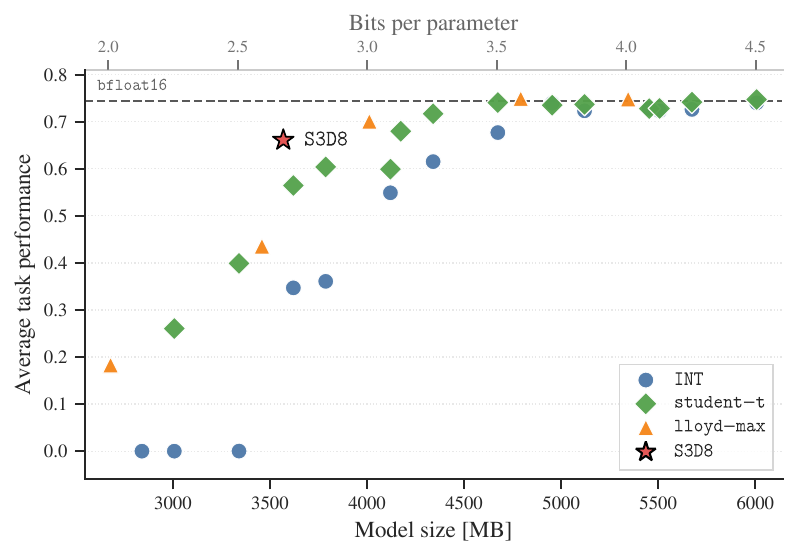}
    \caption{Trade-off between overall task performance and model compression for different numerical formats. Compared to the \texttt{INT} format, \texttt{student-t} and \texttt{lloyd-max} show improved performance-compression trade-off. However, \ourformat{} further improves the trade-off, while allowing for efficient Arm CPU execution.}
    \label{fig:format_overall}
  \end{subfigure}
  \hfill
  \begin{subfigure}[t]{0.49\textwidth}
    \centering
    \includegraphics[width=0.8\linewidth]{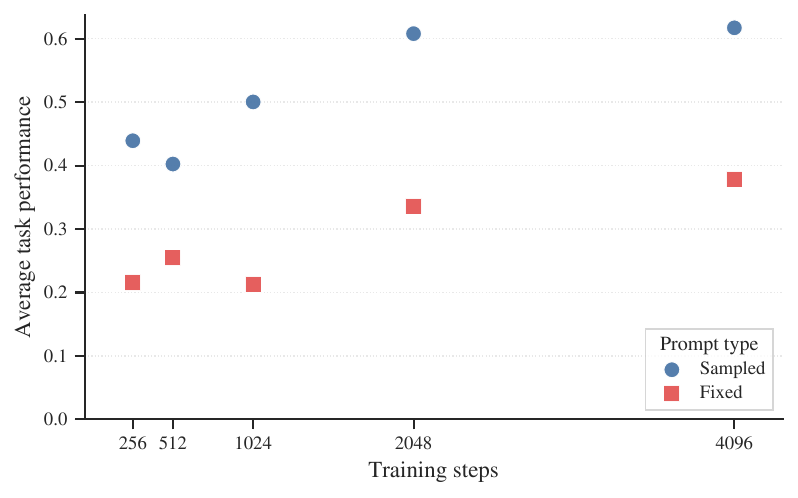}
    \caption{Average task performance vs. QAT training steps for different prompting schemes. \textit{Fixed} prompting uses a single prompt ``Describe the image:'', while \textit{sampled} prompting randomly generates a prompt for each image as described in \cref{sec:qat_data_pipeline}. Even in the knowledge distillation setting, a fixed prompt is too limited in generating diverse outputs, leading to poor task performance.}
    \label{fig:prompt_overall}
  \end{subfigure}

  \caption{Quantization-aware training results.}
  \label{fig:format_prompt_overall}
\end{figure*}


\subsection{Downstream Tasks}
\label{sec:downstream_tasks}

\Cref{tab:format-comparison} shows the downstream performance comparison between the original model \texttt{bfloat16} and our proposed format \ourformat{}, as well as variants quantized with \texttt{INT}, \texttt{student-t}, and \texttt{lloyd-max} using QAT at a similar size. Overall, our format at $2.68$ bits per parameter leads to an average task degradation of $0.083$, significantly outperforming other size-matched formats. The full performance-compression trade-off for these formats is shown in \cref{fig:format_overall}.

\Cref{fig:overview} compares direct casting, GPTQ, and QAT across a range of model sizes. GPTQ improves over direct casting throughout most of the sweep, but QAT remains stronger at lower rates. At approximately $2.7$ bits per parameter, \ourformat{} with GPTQ achieves an average task performance of $0.340$, compared with $0.018$ for rate-matched \texttt{INT} with GPTQ. Using the same quantization method for both isolates the effect of the weight format, showing that \ourformat{} retains substantially more accuracy than \texttt{INT}. Applying QAT to \ourformat{} further increases the average task performance to $0.661$. Full GPTQ settings and implementation details are given in \cref{sec:gptq}.

\subsection{Sampled vs. fixed prompts}
\label{sec:sampled_fixed_prompts}


To measure the effect of prompt sampling (\cref{sec:qat_data_pipeline}), we train a fixed block-scaled \texttt{INT3} format using either sampled prompts or a single fixed prompt (\texttt{``Describe the image:''}), while varying the number of QAT steps (\cref{fig:prompt_overall}). Sampled prompting substantially improves downstream performance, suggesting that QAT benefits from diverse responses that probe a range of instruction-following behavior.

\subsection{Runtime Performance}

Our benchmarking analysis shows the practical efficiency of \ourformat{} as a weight storage format on Arm CPUs. We test on the Android Pixel 8a in two modes: $5$-core and $1$-core, and on a Graviton4 CPU consisting of $96$ Arm Neoverse V2 cores. Full benchmarking configurations are given in \cref{sec:app-benchmarking}.

First, we evaluate the speed of dequantization from $\ourformat{}\rightarrow\texttt{INT8}$. We perform multiple iterations with different \ourformat{} source tensors and a \textit{shared} destination tensor, to simulate the case where the dequantized tensor can remain in on-chip cache. We therefore report read bandwidth (not counting writes), and compare this against an $\texttt{INT8}$ copy with the same constraints --- an upper bound on achievable bandwidth.
Results in \cref{fig:bench-android-dequant,fig:bench-graviton-dequant} show the low overhead of dequantization. For example, for the vision MLP up-projection on the Pixel 8a using $5$ cores, \ourformat{} dequantization takes $133 \upmu$s ($16.5$ GB/s), while the \texttt{INT8} copy takes $310 \upmu$s ($21.2$ GB/s).

Second, we evaluate the speed of dequantize-multiply, with a specific focus on the operation shapes of VLM inference. We report operation shapes as $(m,k,n)$ for $Y = X W^{\top} \in \mathbb{R}^{m \times n}$ from $X \in \mathbb{R}^{m \times k}$. We consider a vision encoder prefill of a single image tile, implying batch size $m\!=\!1601$ patches, followed by a short language prefill of $m\!=\!128$ tokens, then autoregressive generation with batch size $m\!=\!1$. \ourformat{} accelerates most tested generation shapes (small $m$) relative to \texttt{INT8} by reducing memory traffic. For larger $m$, dequantization adds work to the compute-bound \texttt{INT8} multiply. On Graviton4, \ourformat{} incurs a modest prefill slowdown relative to \texttt{INT8}; on the five-core Pixel, prefill performance is similar (\cref{tbl:bench-matmul-summary}).

\begin{table}
    \caption{Compute rate of selected matrix multiply shapes across different weight formats.
    \ourformat{} uses a fused kernel for $m\!=\!1$ and a cast followed by \texttt{INT8} multiply for $m\!>\!1$.
    See \cref{tbl:bench-matmul-android-mt,tbl:bench-matmul-android-1t,tbl:bench-matmul-graviton} for further results.
    }
    \label{tbl:bench-matmul-summary}
    \centering
\begin{tabular}{p{2.5cm} C{1.4cm} C{1.4cm} C{1.4cm}}
\toprule
Shape & \multicolumn{3}{c}{GMAC/s} \\
$(m,k,n)$ & $\mathtt{bfloat16}$ & $\mathtt{INT8}$ & $\mathtt{S3D8}$ \\
\midrule
\multicolumn{3}{l}{\textit{Pixel 8a $5$-core}} \\
$(1, 4096, 14336)$ & $13.6$ & $26.5$ & $\mathbf{33.8}$ \\
$(128, 4096, 14336)$ & $76.3$ & $\mathbf{201.1}$ & $\mathbf{197.7}$ \\
$(1601, 1280, 5120)$ & $116.5$ & $\mathbf{206.1}$ & $\mathbf{211.8}$ \\
\midrule
\multicolumn{3}{l}{\textit{Graviton4}} \\
$(1, 4096, 14336)$ & $236$ & $461$ & $\mathbf{1031}$ \\
$(128, 4096, 14336)$ & $3492$ & $\mathbf{7700}$ & $6994$ \\
$(1601, 1280, 5120)$ & $4210$ & $\mathbf{8464}$ & $7979$ \\
\bottomrule
\end{tabular}
\end{table}

\subsection{End-to-end Deployment on Arm CPUs}

We constructed a custom C++ inference implementation of the Llama 3.2 11B Vision model. It includes a custom file format based on \texttt{safetensors} \cite{safetensors} and efficient fused kernel implementations. We used this implementation to validate end-to-end support using \ourformat{} from a model file of size $3.73$ GB, including the vocabulary and metadata.

To evaluate end-to-end performance, we generate $100$ tokens from a single image tile and short prompt and measure the token/s for the decoding phase, excluding model loading and image/text prefill. Our implementation achieves a median $3.8$ tokens/s ($12.5$ GB/s parameter read bandwidth) with \ourformat{} on the Pixel 8a using $5$ OpenMP threads on the prime and performance cores --- an appreciable fraction of the maximum observed read bandwidth $\approx 25$ GB/s. With parameters in \texttt{INT8}, the model does not fit in memory. On Graviton4, we obtain $36.8$ tokens/s ($120$ GB/s) for \ourformat{} versus $26.4$ tokens/s ($258$ GB/s) for \texttt{INT8}, showing a practical speedup for single-user text generation.

\section{Limitations}

While our quantization framework is general, our experiments focus on a single model, Llama 3.2 11B Vision Instruct, and on CPU execution. \ourformat{} uses an Arm-optimized $64$-entry table lookup decoder; other architectures may require specialized kernels. Our evaluation also uses a small set of VQA-style benchmarks that were part of the original Llama evaluation suite. We used a fixed quantization format across all model weights; preliminary experiments with keeping selected layers in higher precision did not improve the accuracy-compression trade-off. Our comparisons use our own implementations of the quantization methods and formats; accuracy and runtime relative to deployed formats such as \texttt{llama.cpp}'s \texttt{Q2\_K} and \texttt{IQ2} families remain unevaluated. Finally, the prompt and image selection may be further optimized to improve downstream task generalization.

\section{Conclusion}

We presented Llama-Mobile, a framework for aggressively quantizing vision-language models for low-resource inference. Our approach combines a synthetic multimodal data pipeline for quantization-aware distillation and \ourformat{}, a $2.7$-bit-per-parameter weight format designed for efficient Arm CPU execution. Applied to Llama 3.2 11B Vision Instruct, the method compresses the model to $3.7$ GB with $8$-bit activations, while preserving good performance across a set of VQA benchmarks. Our runtime performance analysis supports the practical efficiency of our method. Together, these results suggest that sub-$3$-bit VLM compression is a viable route toward multimodal inference on resource-constrained devices.

\subsection{Acknowledgments}

We thank Eric Biscondi, Dominic Pajak, Dobrik Georgiev, Guoxuan Xia, Tom Cashman, Paul Balança and Carlo Luschi for their helpful advice and feedback on this work.

\printbibliography

@article{llama3,
  author       = {{Llama Team}},
  title        = {The Llama 3 Herd of Models},
  journal      = {CoRR},
  volume       = {abs/2407.21783},
  year         = {2024},
  commentedurl          = {https://doi.org/10.48550/arXiv.2407.21783},
  doi          = {10.48550/ARXIV.2407.21783},
  eprinttype   = {arXiv},
  eprint       = {2407.21783},
  bibsource    = {dblp computer science bibliography, https://dblp.org}
}

@article{gemma3,
  author       = {{Gemma Team}},
  title        = {Gemma 3 Technical Report},
  journal      = {CoRR},
  volume       = {abs/2503.19786},
  year         = {2025},
  commentedurl          = {https://doi.org/10.48550/arXiv.2503.19786},
  doi          = {10.48550/ARXIV.2503.19786},
  eprinttype   = {arXiv},
  eprint       = {2503.19786},
  bibsource    = {dblp computer science bibliography, https://dblp.org}
}

@article{vqav2,
  author       = {Yash Goyal and
                  Tejas Khot and
                  Aishwarya Agrawal and
                  Douglas Summers{-}Stay and
                  Dhruv Batra and
                  Devi Parikh},
  title        = {Making the {V} in {VQA} Matter: Elevating the Role of Image Understanding
                  in Visual Question Answering},
  journal      = {Int. J. Comput. Vis.},
  volume       = {127},
  number       = {4},
  pages        = {398--414},
  year         = {2019},
  commentedurl          = {https://doi.org/10.1007/s11263-018-1116-0},
  doi          = {10.1007/S11263-018-1116-0},
  bibsource    = {dblp computer science bibliography, https://dblp.org}
}

@inproceedings{docvqa,
  author       = {Minesh Mathew and
                  Dimosthenis Karatzas and
                  C. V. Jawahar},
  title        = {DocVQA: {A} Dataset for {VQA} on Document Images},
  booktitle    = {{IEEE} Winter Conference on Applications of Computer Vision, {WACV}
                  2021, Waikoloa, HI, USA, January 3-8, 2021},
  pages        = {2199--2208},
  publisher    = {{IEEE}},
  year         = {2021},
  commentedurl          = {https://doi.org/10.1109/WACV48630.2021.00225},
  doi          = {10.1109/WACV48630.2021.00225},
  bibsource    = {dblp computer science bibliography, https://dblp.org}
}

@inproceedings{chartqa,
  author       = {Ahmed Masry and
                  Do Xuan Long and
                  Jia Qing Tan and
                  Shafiq R. Joty and
                  Enamul Hoque},
  editor       = {Smaranda Muresan and
                  Preslav Nakov and
                  Aline Villavicencio},
  title        = {ChartQA: {A} Benchmark for Question Answering about Charts with Visual
                  and Logical Reasoning},
  booktitle    = {Findings of the Association for Computational Linguistics: {ACL} 2022,
                  Dublin, Ireland, May 22-27, 2022},
  series       = {Findings of {ACL}},
  pages        = {2263--2279},
  publisher    = {Association for Computational Linguistics},
  year         = {2022},
  commentedurl          = {https://doi.org/10.18653/v1/2022.findings-acl.177},
  doi          = {10.18653/V1/2022.FINDINGS-ACL.177},
  bibsource    = {dblp computer science bibliography, https://dblp.org}
}

@inproceedings{ai2d,
  author       = {Aniruddha Kembhavi and
                  Mike Salvato and
                  Eric Kolve and
                  Min Joon Seo and
                  Hannaneh Hajishirzi and
                  Ali Farhadi},
  editor       = {Bastian Leibe and
                  Jiri Matas and
                  Nicu Sebe and
                  Max Welling},
  title        = {A Diagram is Worth a Dozen Images},
  booktitle    = {Computer Vision - {ECCV} 2016 - 14th European Conference, Amsterdam,
                  The Netherlands, October 11-14, 2016, Proceedings, Part {IV}},
  series       = {Lecture Notes in Computer Science},
  pages        = {235--251},
  publisher    = {Springer},
  year         = {2016},
  commentedurl          = {https://doi.org/10.1007/978-3-319-46493-0\_15},
  doi          = {10.1007/978-3-319-46493-0\_15},
  bibsource    = {dblp computer science bibliography, https://dblp.org}
}

@article{gemini2.5,
  author       = {{Gemini Team}},
  title        = {Gemini 2.5: Pushing the Frontier with Advanced Reasoning, Multimodality,
                  Long Context, and Next Generation Agentic Capabilities},
  journal      = {CoRR},
  volume       = {abs/2507.06261},
  year         = {2025},
  commentedurl          = {https://doi.org/10.48550/arXiv.2507.06261},
  doi          = {10.48550/ARXIV.2507.06261},
  eprinttype   = {arXiv},
  eprint       = {2507.06261},
  bibsource    = {dblp computer science bibliography, https://dblp.org}
}

@article{qwen3,
      author={{Qwen Team}},
      title={Qwen3 Technical Report}, 
      journal= {CoRR},
      volume={abs/2505.09388},
      year={2025},
      commentedurl={https://arxiv.org/abs/2505.09388}, 
      eprinttype={arXiv},
      eprint={2505.09388},
}

@article{qwen3vl,
  author       = {{Qwen Team}},
  title        = {Qwen3-VL Technical Report},
  journal      = {CoRR},
  volume       = {abs/2511.21631},
  year         = {2025},
  commentedurl          = {https://doi.org/10.48550/arXiv.2511.21631},
  doi          = {10.48550/ARXIV.2511.21631},
  eprinttype   = {arXiv},
  eprint       = {2511.21631},
  bibsource    = {dblp computer science bibliography, https://dblp.org}
}

@article{deepseek3.2,
  author       = {DeepSeek{-}AI},
  title        = {DeepSeek-V3.2: Pushing the Frontier of Open Large Language Models},
  journal      = {CoRR},
  volume       = {abs/2512.02556},
  year         = {2025},
  commentedurl          = {https://doi.org/10.48550/arXiv.2512.02556},
  doi          = {10.48550/ARXIV.2512.02556},
  eprinttype   = {arXiv},
  eprint       = {2512.02556},
  bibsource    = {dblp computer science bibliography, https://dblp.org}
}

@article{frantar2022gptq,
  author       = {Elias Frantar and
                  Saleh Ashkboos and
                  Torsten Hoefler and
                  Dan Alistarh},
  title        = {{GPTQ:} Accurate Post-Training Quantization for Generative Pre-trained
                  Transformers},
  journal      = {CoRR},
  volume       = {abs/2210.17323},
  year         = {2022},
  commentedurl          = {https://doi.org/10.48550/arXiv.2210.17323},
  doi          = {10.48550/ARXIV.2210.17323},
  eprinttype   = {arXiv},
  eprint       = {2210.17323},
  bibsource    = {dblp computer science bibliography, https://dblp.org}
}

@inproceedings{liu2024llmqat,
  author       = {Zechun Liu and
                  Barlas Oguz and
                  Changsheng Zhao and
                  Ernie Chang and
                  Pierre Stock and
                  Yashar Mehdad and
                  Yangyang Shi and
                  Raghuraman Krishnamoorthi and
                  Vikas Chandra},
  editor       = {Lun{-}Wei Ku and
                  Andre Martins and
                  Vivek Srikumar},
  title        = {{LLM-QAT:} Data-Free Quantization Aware Training for Large Language
                  Models},
  booktitle    = {Findings of the Association for Computational Linguistics, {ACL} 2024,
                  Bangkok, Thailand and virtual meeting, August 11-16, 2024},
  series       = {Findings of {ACL}},
  pages        = {467--484},
  publisher    = {Association for Computational Linguistics},
  year         = {2024},
  commentedurl          = {https://doi.org/10.18653/v1/2024.findings-acl.26},
  doi          = {10.18653/V1/2024.FINDINGS-ACL.26},
  bibsource    = {dblp computer science bibliography, https://dblp.org}
}

@article{orr2025optimal,
  author       = {Douglas Orr and
                  Luka Ribar and
                  Carlo Luschi},
  title        = {Optimal Formats for Weight Quantisation},
  journal      = {CoRR},
  volume       = {abs/2505.12988},
  year         = {2025},
  commentedurl          = {https://doi.org/10.48550/arXiv.2505.12988},
  doi          = {10.48550/ARXIV.2505.12988},
  eprinttype   = {arXiv},
  eprint       = {2505.12988},
  bibsource    = {dblp computer science bibliography, https://dblp.org}
}

@inproceedings{imagenet,
  author       = {Jia Deng and
                  Wei Dong and
                  Richard Socher and
                  Li{-}Jia Li and
                  Kai Li and
                  Li Fei{-}Fei},
  title        = {ImageNet: {A} large-scale hierarchical image database},
  booktitle    = {2009 {IEEE} Computer Society Conference on Computer Vision and Pattern
                  Recognition {(CVPR} 2009), 20-25 June 2009, Miami, Florida, {USA}},
  pages        = {248--255},
  publisher    = {{IEEE} Computer Society},
  year         = {2009},
  commentedurl          = {https://doi.org/10.1109/CVPR.2009.5206848},
  doi          = {10.1109/CVPR.2009.5206848},
  bibsource    = {dblp computer science bibliography, https://dblp.org}
}

@article{bengio2013ste,
  author       = {Yoshua Bengio and
                  Nicholas L{\'{e}}onard and
                  Aaron C. Courville},
  title        = {Estimating or Propagating Gradients Through Stochastic Neurons for
                  Conditional Computation},
  journal      = {CoRR},
  volume       = {abs/1308.3432},
  year         = {2013},
  commentedurl          = {http://arxiv.org/abs/1308.3432},
  eprinttype   = {arXiv},
  eprint       = {1308.3432},
  bibsource    = {dblp computer science bibliography, https://dblp.org}
}

@misc{llama_vision_model_card,
  author  = {{Meta}},
  title   = {{Llama 3.2-Vision Model Card}},
  year    = {2024},
  date    = {2024-09-25},
  url     = {https://github.com/meta-llama/llama-models/blob/main/models/llama3_2/MODEL_CARD_VISION.md},
  urldate = {2026-04-28}
}

@article{lloyd1982kmeans,
  author       = {Stuart P. Lloyd},
  title        = {Least squares quantization in {PCM}},
  journal      = {{IEEE} Trans. Inf. Theory},
  volume       = {28},
  number       = {2},
  pages        = {129--136},
  year         = {1982},
  commentedurl          = {https://doi.org/10.1109/TIT.1982.1056489},
  doi          = {10.1109/TIT.1982.1056489},
  bibsource    = {dblp computer science bibliography, https://dblp.org}
}

@inproceedings{arthur2007kmeans++,
  author       = {David Arthur and
                  Sergei Vassilvitskii},
  editor       = {Nikhil Bansal and
                  Kirk Pruhs and
                  Clifford Stein},
  title        = {k-means++: the advantages of careful seeding},
  booktitle    = {Proceedings of the Eighteenth Annual {ACM-SIAM} Symposium on Discrete
                  Algorithms, {SODA} 2007, New Orleans, Louisiana, USA, January 7-9,
                  2007},
  pages        = {1027--1035},
  publisher    = {{SIAM}},
  year         = {2007},
  commentedurl          = {http://dl.acm.org/citation.cfm?id=1283383.1283494},
  bibsource    = {dblp computer science bibliography, https://dblp.org}
}

@misc{openmp,
  title        = {{OpenMP} Application Programming Interface Version 5.1},
  author       = {{OpenMP Architecture Review Board}},
  year         = {2020},
  url          = {https://www.openmp.org/specifications/},
  urldate      = {2026-04-28}
}

@article{efron1979bootstrap,
  title   = {Bootstrap Methods: Another Look at the Jackknife},
  author  = {Bradley Efron},
  journal = {The Annals of Statistics},
  year    = {1979},
  volume  = {7},
  number  = {1},
  pages   = {1--26},
  doi     = {10.1214/aos/1176344552}
}

@misc{armarch,
  title        = {{Arm Architecture Reference Manual for A-profile Architecture}},
  author       = {{Arm Ltd.}},
  year         = {2026},
  url          = {https://developer.arm.com/documentation/ddi0487/latest},
  urldate      = {2026-04-28}
}

@misc{safetensors,
  title        = {{Safetensors}: Simple, Safe Tensor Serialization},
  author       = {{Hugging Face}},
  year         = {2026},
  url          = {https://github.com/huggingface/safetensors},
  urldate      = {2026-04-30}
}

@inproceedings{dettmers2023scaling,
    author = {Dettmers, Tim and Zettlemoyer, Luke},
    title = {The case for 4-bit precision: k-bit inference scaling laws},
    year = {2023},
    publisher = {JMLR.org},
    booktitle = {Proceedings of the 40th International Conference on Machine Learning},
    articleno = {307},
    numpages = {25},
    location = {Honolulu, Hawaii, USA},
    series = {ICML'23}
}

@inproceedings{xiao2023smoothquant,
  author       = {Guangxuan Xiao and
                  Ji Lin and
                  Micka{\"{e}}l Seznec and
                  Hao Wu and
                  Julien Demouth and
                  Song Han},
  editor       = {Andreas Krause and
                  Emma Brunskill and
                  Kyunghyun Cho and
                  Barbara Engelhardt and
                  Sivan Sabato and
                  Jonathan Scarlett},
  title        = {SmoothQuant: Accurate and Efficient Post-Training Quantization for
                  Large Language Models},
  booktitle    = {International Conference on Machine Learning, {ICML} 2023, 23-29 July
                  2023, Honolulu, Hawaii, {USA}},
  series       = {Proceedings of Machine Learning Research},
  pages        = {38087--38099},
  publisher    = {{PMLR}},
  year         = {2023},
  commentedurl          = {https://proceedings.mlr.press/v202/xiao23c.html},
  bibsource    = {dblp computer science bibliography, https://dblp.org}
}

@article{deepseek2024,
  author       = {DeepSeek{-}AI},
  title        = {DeepSeek-V3 Technical Report},
  journal      = {CoRR},
  volume       = {abs/2412.19437},
  year         = {2024},
  commentedurl          = {https://doi.org/10.48550/arXiv.2412.19437},
  doi          = {10.48550/ARXIV.2412.19437},
  eprinttype   = {arXiv},
  eprint       = {2412.19437},
  bibsource    = {dblp computer science bibliography, https://dblp.org}
}

@article{huang2026sherry,
  author       = {Hong Huang and
                  Decheng Wu and
                  Qiangqiang Hu and
                  Guanghua Yu and
                  Jinhai Yang and
                  Jianchen Zhu and
                  Xue Liu and
                  Dapeng Wu},
  title        = {Sherry: Hardware-Efficient 1.25-Bit Ternary Quantization via Fine-grained
                  Sparsification},
  journal      = {CoRR},
  volume       = {abs/2601.07892},
  year         = {2026},
  commentedurl          = {https://doi.org/10.48550/arXiv.2601.07892},
  doi          = {10.48550/ARXIV.2601.07892},
  eprinttype   = {arXiv},
  eprint       = {2601.07892},
  bibsource    = {dblp computer science bibliography, https://dblp.org}
}

@inproceedings{egiazarian2024aqlm,
    author = {Egiazarian, Vage and Panferov, Andrei and Kuznedelev, Denis and Frantar, Elias and Babenko, Artem and Alistarh, Dan},
    title = {Extreme compression of large language models via additive quantization},
    year = {2024},
    publisher = {JMLR.org},
    booktitle = {Proceedings of the 41st International Conference on Machine Learning},
    articleno = {488},
    numpages = {20},
    location = {Vienna, Austria},
    series = {ICML'24}
}

@misc{ocpspec,
  title        = {{OCP} Microscaling Formats {(MX)} Specification v1.0},
  author       = {{OCP}},
  year         = {2023},
  url          = {https://www.opencompute.org/documents/ocp-microscaling-formats-mx-v1-0-spec-final-pdf},
  urldate      = {2026-04-30}
}

@article{rouhani2023microscaling,
  author       = {Bita Darvish Rouhani and
                  Ritchie Zhao and
                  Ankit More and
                  Mathew Hall and
                  Alireza Khodamoradi and
                  Summer Deng and
                  Dhruv Choudhary and
                  Marius Cornea and
                  Eric Dellinger and
                  Kristof Denolf and
                  Dusan Stosic and
                  Venmugil Elango and
                  Maximilian Golub and
                  Alexander Heinecke and
                  Phil James{-}Roxby and
                  Dharmesh Jani and
                  Gaurav Kolhe and
                  Martin Langhammer and
                  Ada Li and
                  Levi Melnick and
                  Maral Mesmakhosroshahi and
                  Andres Rodriguez and
                  Michael Schulte and
                  Rasoul Shafipour and
                  Lei Shao and
                  Michael Y. Siu and
                  Pradeep Dubey and
                  Paulius Micikevicius and
                  Maxim Naumov and
                  Colin Verilli and
                  Ralph Wittig and
                  Doug Burger and
                  Eric S. Chung},
  title        = {Microscaling Data Formats for Deep Learning},
  journal      = {CoRR},
  volume       = {abs/2310.10537},
  year         = {2023},
  commentedurl          = {https://doi.org/10.48550/arXiv.2310.10537},
  doi          = {10.48550/ARXIV.2310.10537},
  eprinttype   = {arXiv},
  eprint       = {2310.10537},
  bibsource    = {dblp computer science bibliography, https://dblp.org}
}

@article{liu2025paretoq,
  author       = {Zechun Liu and
                  Changsheng Zhao and
                  Hanxian Huang and
                  Sijia Chen and
                  Jing Zhang and
                  Jiawei Zhao and
                  Scott Roy and
                  Lisa Jin and
                  Yunyang Xiong and
                  Yangyang Shi and
                  Lin Xiao and
                  Yuandong Tian and
                  Bilge Soran and
                  Raghuraman Krishnamoorthi and
                  Tijmen Blankevoort and
                  Vikas Chandra},
  title        = {ParetoQ: Scaling Laws in Extremely Low-bit {LLM} Quantization},
  journal      = {CoRR},
  volume       = {abs/2502.02631},
  year         = {2025},
  commentedurl          = {https://doi.org/10.48550/arXiv.2502.02631},
  doi          = {10.48550/ARXIV.2502.02631},
  eprinttype   = {arXiv},
  eprint       = {2502.02631},
  bibsource    = {dblp computer science bibliography, https://dblp.org}
}

@inproceedings{dettmers2023qlora,
  author       = {Tim Dettmers and
                  Artidoro Pagnoni and
                  Ari Holtzman and
                  Luke Zettlemoyer},
  editor       = {Alice Oh and
                  Tristan Naumann and
                  Amir Globerson and
                  Kate Saenko and
                  Moritz Hardt and
                  Sergey Levine},
  title        = {QLoRA: Efficient Finetuning of Quantized LLMs},
  booktitle    = {Advances in Neural Information Processing Systems 36: Annual Conference
                  on Neural Information Processing Systems 2023, NeurIPS 2023, New Orleans,
                  LA, USA, December 10 - 16, 2023},
  year         = {2023},
  commentedurl          = {http://papers.nips.cc/paper\_files/paper/2023/hash/1feb87871436031bdc0f2beaa62a049b-Abstract-Conference.html},
  bibsource    = {dblp computer science bibliography, https://dblp.org}
}

@inproceedings{wang2024qvlm,
  author       = {Changyuan Wang and
                  Ziwei Wang and
                  Xiuwei Xu and
                  Yansong Tang and
                  Jie Zhou and
                  Jiwen Lu},
  editor       = {Amir Globersons and
                  Lester Mackey and
                  Danielle Belgrave and
                  Angela Fan and
                  Ulrich Paquet and
                  Jakub M. Tomczak and
                  Cheng Zhang},
  title        = {{Q-VLM:} Post-training Quantization for Large Vision-Language Models},
  booktitle    = {Advances in Neural Information Processing Systems 38: Annual Conference
                  on Neural Information Processing Systems 2024, NeurIPS 2024, Vancouver,
                  BC, Canada, December 10 - 15, 2024},
  year         = {2024},
  commentedurl          = {http://papers.nips.cc/paper\_files/paper/2024/hash/cffbaf4f47546ece96bb42c0edda40ee-Abstract-Conference.html},
  bibsource    = {dblp computer science bibliography, https://dblp.org}
}

@inproceedings{li2025mbq,
  author       = {Shiyao Li and
                  Yingchun Hu and
                  Xuefei Ning and
                  Xihui Liu and
                  Ke Hong and
                  Xiaotao Jia and
                  Xiuhong Li and
                  Yaqi Yan and
                  Pei Ran and
                  Guohao Dai and
                  Shengen Yan and
                  Huazhong Yang and
                  Yu Wang},
  title        = {{MBQ:} Modality-Balanced Quantization for Large Vision-Language Models},
  booktitle    = {{IEEE/CVF} Conference on Computer Vision and Pattern Recognition,
                  {CVPR} 2025, Nashville, TN, USA, June 11-15, 2025},
  pages        = {4167--4177},
  publisher    = {Computer Vision Foundation / {IEEE}},
  year         = {2025},
  commentedurl          = {https://openaccess.thecvf.com/content/CVPR2025/html/Li\_MBQ\_Modality-Balanced\_Quantization\_for\_Large\_Vision-Language\_Models\_CVPR\_2025\_paper.html},
  doi          = {10.1109/CVPR52734.2025.00394},
  bibsource    = {dblp computer science bibliography, https://dblp.org}
}

@article{xue2025vlmq,
  author       = {Yufei Xue and
                  Yushi Huang and
                  Jiawei Shao and
                  Jun Zhang},
  title        = {{VLMQ:} Efficient Post-Training Quantization for Large Vision-Language
                  Models via Hessian Augmentation},
  journal      = {CoRR},
  volume       = {abs/2508.03351},
  year         = {2025},
  commentedurl          = {https://doi.org/10.48550/arXiv.2508.03351},
  doi          = {10.48550/ARXIV.2508.03351},
  eprinttype   = {arXiv},
  eprint       = {2508.03351},
  bibsource    = {dblp computer science bibliography, https://dblp.org}
}

@article{gope2025kernels,
  author       = {Dibakar Gope and
                  David Mansell and
                  Danny Loh and
                  Ian Bratt},
  title        = {Highly Optimized Kernels and Fine-Grained Codebooks for {LLM} Inference
                  on Arm CPUs},
  journal      = {CoRR},
  volume       = {abs/2501.00032},
  year         = {2025},
  commentedurl          = {https://doi.org/10.48550/arXiv.2501.00032},
  doi          = {10.48550/ARXIV.2501.00032},
  eprinttype   = {arXiv},
  eprint       = {2501.00032},
  bibsource    = {dblp computer science bibliography, https://dblp.org}
}

@inproceedings{son2025nsnquant,
  author       = {Donghyun Son and
                  Euntae Choi and
                  Sungjoo Yoo},
  editor       = {Danielle Belgrave and
                  Cheng Zhang and
                  Laura N. Montoya and
                  Hsuan{-}Tien Lin and
                  Razvan Pascanu and
                  Piotr Koniusz and
                  Marzyeh Ghassemi and
                  Nancy Chen and
                  Iv{\'{a}}n Vladimir Meza Ru{\'{\i}}z and
                  Arturo Loaiza{-}Bonilla},
  title        = {NSNQuant: {A} Double Normalization Approach for Calibration-Free Low-Bit
                  Vector Quantization of {KV} Cache},
  booktitle    = {Advances in Neural Information Processing Systems 38: Annual Conference
                  on Neural Information Processing Systems 2025, NeurIPS 2025, San Diego,
                  CA, USA, December 2-7, 2025 / Mexico City, Mexico, November 30 - December
                  5, 2025},
  year         = {2025},
  url          = {http://papers.nips.cc/paper\_files/paper/2025/hash/3d8ee933c215fbb7b4d1948ff4906299-Abstract-Conference.html},
  bibsource    = {dblp computer science bibliography, https://dblp.org}
}

@misc{llamacpp,
  title        = {{llama.cpp}: {Port of {LLaMA} Model in {C/C++}}},
  author       = {Georgi Gerganov},
  year         = {2023},
  url          = {https://github.com/ggml-org/llama.cpp},
  urldate      = {2026-05-01}
}

@article{xin2026qad-nvfp4,
  author       = {Meng Xin and
                  Sweta Priyadarshi and
                  Jingyu Xin and
                  Bilal Kartal and
                  Aditya Vavre and
                  Asma Kuriparambil Thekkumpate and
                  Zijia Chen and
                  Ameya Sunil Mahabaleshwarkar and
                  Ido Shahaf and
                  Akhiad Bercovich and
                  Kinjal Patel and
                  Suguna Varshini Velury and
                  Chenjie Luo and
                  Zhiyu Cheng and
                  Jenny Chen and
                  Chen{-}Han Yu and
                  Wei Ping and
                  Oleg Rybakov and
                  Nima Tajbakhsh and
                  Oluwatobi Olabiyi and
                  Dusan Stosic and
                  Di Wu and
                  Song Han and
                  Eric Chung and
                  Sharath Turuvekere Sreenivas and
                  Bryan Catanzaro and
                  Yoshi Suhara and
                  Tijmen Blankevoort and
                  Huizi Mao},
  title        = {Quantization-Aware Distillation for {NVFP4} Inference Accuracy Recovery},
  journal      = {CoRR},
  volume       = {abs/2601.20088},
  year         = {2026},
  commentedurl          = {https://doi.org/10.48550/arXiv.2601.20088},
  doi          = {10.48550/ARXIV.2601.20088},
  eprinttype   = {arXiv},
  eprint       = {2601.20088},
  bibsource    = {dblp computer science bibliography, https://dblp.org}
}

@inproceedings{mathew2022infographicvqa,
  author       = {Minesh Mathew and
                  Viraj Bagal and
                  Rub{\`{e}}n Tito and
                  Dimosthenis Karatzas and
                  Ernest Valveny and
                  C. V. Jawahar},
  title        = {InfographicVQA},
  booktitle    = {{IEEE/CVF} Winter Conference on Applications of Computer Vision, {WACV}
                  2022, Waikoloa, HI, USA, January 3-8, 2022},
  pages        = {2582--2591},
  publisher    = {{IEEE}},
  year         = {2022},
  commentedurl          = {https://doi.org/10.1109/WACV51458.2022.00264},
  doi          = {10.1109/WACV51458.2022.00264},
  bibsource    = {dblp computer science bibliography, https://dblp.org}
}

@inproceedings{du2025optimize_quantization,
  author       = {Jiangcun Du and
                  Renren Jin and
                  Wuwei Huang and
                  Wei Liu and
                  Jian Luan and
                  Deyi Xiong},
  editor       = {Guodong Long and
                  Michale Blumestein and
                  Yi Chang and
                  Liane Lewin{-}Eytan and
                  Zi Helen Huang and
                  Elad Yom{-}Tov},
  title        = {Optimize Quantization for Large Language Models via Progressive Training},
  booktitle    = {Companion Proceedings of the {ACM} on Web Conference 2025, {WWW} 2025,
                  Sydney, NSW, Australia, 28 April 2025 - 2 May 2025},
  pages        = {2474--2483},
  publisher    = {{ACM}},
  year         = {2025},
  commentedurl          = {https://doi.org/10.1145/3701716.3717578},
  doi          = {10.1145/3701716.3717578},
  bibsource    = {dblp computer science bibliography, https://dblp.org}
}

@inproceedings{liu2023llava,
  author       = {Haotian Liu and
                  Chunyuan Li and
                  Qingyang Wu and
                  Yong Jae Lee},
  editor       = {Alice Oh and
                  Tristan Naumann and
                  Amir Globerson and
                  Kate Saenko and
                  Moritz Hardt and
                  Sergey Levine},
  title        = {Visual Instruction Tuning},
  booktitle    = {Advances in Neural Information Processing Systems 36: Annual Conference
                  on Neural Information Processing Systems 2023, NeurIPS 2023, New Orleans,
                  LA, USA, December 10 - 16, 2023},
  year         = {2023},
  commentedurl          = {http://papers.nips.cc/paper\_files/paper/2023/hash/6dcf277ea32ce3288914faf369fe6de0-Abstract-Conference.html},
  bibsource    = {dblp computer science bibliography, https://dblp.org}
}

@inproceedings{dai2023instructblip,
  author       = {Wenliang Dai and
                  Junnan Li and
                  Dongxu Li and
                  Anthony Meng Huat Tiong and
                  Junqi Zhao and
                  Weisheng Wang and
                  Boyang Li and
                  Pascale Fung and
                  Steven C. H. Hoi},
  editor       = {Alice Oh and
                  Tristan Naumann and
                  Amir Globerson and
                  Kate Saenko and
                  Moritz Hardt and
                  Sergey Levine},
  title        = {InstructBLIP: Towards General-purpose Vision-Language Models with
                  Instruction Tuning},
  booktitle    = {Advances in Neural Information Processing Systems 36: Annual Conference
                  on Neural Information Processing Systems 2023, NeurIPS 2023, New Orleans,
                  LA, USA, December 10 - 16, 2023},
  year         = {2023},
  commentedurl          = {http://papers.nips.cc/paper\_files/paper/2023/hash/9a6a435e75419a836fe47ab6793623e6-Abstract-Conference.html},
  bibsource    = {dblp computer science bibliography, https://dblp.org}
}

\clearpage
\appendix

\section{Experimental Details}
\label{app:experimental_details}

\subsection{Data Generation}
\label{sec:data_generation}
For the image database, we use the ImageNet \cite{imagenet} training set, which contains $1{,}281{,}167$ images in total. To generate the image-text pairs for QAT, we sampled a random prompt for each image and passed the image-prompt pair as inputs to the Llama 3.2 11B Vision Instruct model in \texttt{bfloat16}. The responses were sampled with temperature $0.6$ and $\mathrm{top\_p}=0.9$, with the maximum number of generated tokens set to $512$, and each response was saved as text. The procedure leads to variable sequence length of the generated text, as the model can stop early once it generates the special \texttt{<|eot\_id|>} token. This is desirable as it leads to the training data containing variable length responses based on the prompt, thus improving the diversity of the saved responses.

For each sample, we generate the random prompt in the following way. In order to induce the instruction-tuned behavior, the Llama 3.2 11B Vision Instruct model uses the following template:
\begin{lstlisting}[style=promptstyle]
<|start_header_id|>user<|end_header_id|>\n\n<|image|>{prompt}<|eot_id|><|start_header_id|>assistant<|end_header_id|>\n\n
\end{lstlisting}
In order to keep the instruction-following behavior of the original model without breaking its ability to generate in the non-instruction setting, we apply the template with probability $p_{\mathrm{temp}}=0.75$. With probability $1-p_{\mathrm{temp}}$ we instead use the plain template:
\begin{lstlisting}[style=promptstyle]
<|image|>{prompt}
\end{lstlisting}

Next, we sample the prompt. First we sample the ``question'' part of the prompt, which is a generic image-comprehension question that can be applied to most natural images. The question pool combines hand-written generic prompts, such as \texttt{"Describe the image."} or \texttt{"Summarize the scene."}, with generated prompts of the form \verb|{action} {target} {audience} {style}|.

For each generated prompt, we sample an action, such as \texttt{"Describe"} or \texttt{"Explain"}, a target, such as \texttt{"the main subject"} or \texttt{"the key objects and their relationships"}, an audience, such as \texttt{"for a dataset annotation"} or \texttt{"for a quick human review"}, and a style, such as \texttt{"using plain language"} or \texttt{"while explicitly noting uncertainty"}.

This gives a pool of $495$ question prompts before instruction blocks are added.

Once we generate the prompt, we prepend an additional instruction with probability $p_{\mathrm{inst}} = 0.7$. The instruction block always contains a generic image-reasoning instruction, such as \texttt{"Use simple language and avoid jargon"}. It may also contain an adherence probe, length probe, or format instruction, sampled independently with probabilities $p_{\mathrm{adh}}=0.4$, $p_{\mathrm{len}}=0.75$, and $p_{\mathrm{fmt}}=0.45$, respectively.

When a length probe is used, short, medium, and long response constraints are selected with probabilities $0.40$, $0.38$, and $0.22$, respectively. This biases the generated data toward concise responses while still including some longer visual explanations.

Finally, for the ``fixed prompt'' experiments, we used the same \verb|"Describe the image:\n"| prompt for each image.

In all cases, no downstream benchmark images or benchmark-specific prompts are used during synthetic data generation.

\subsection{Training}
All of the training examples for the QAT were constructed by randomly sampling images from the ImageNet train set, combined with the random prompts as described in \cref{sec:data_generation}. All QAT results used batch size $128$, with a maximum sequence length $512$. All quantized models were trained for $2048$ steps, apart from the sweep results in \cref{fig:prompt_overall}. We used AdamW optimizer, with $\beta_1 = 0.9$ and $\beta_2 = 0.95$, and a cosine learning schedule. We set the learning rate as $\eta = 2^{- (b + 14)}$, where $b$ is the average number of bits per parameter, following the scaling heuristic from \textcite{orr2025optimal}. During training, master weights were kept in \texttt{float32} and the quantization function was applied in the forward pass. At each training step, we tokenize the saved teacher generations, compute teacher and student logits on the generated answer tokens, and use the full logits vectors for the KL divergence.

\subsection{Quantization Formats}
\label{sec:quant-formats}

The scalar baseline formats use $K$ codepoints, with element bit-width $b=\log_2 K$. For each baseline we sweep $K \in \{4,6,8,12,16\}$ and train for $2048$ QAT steps. Block and channel scaling use absmax scaling with \texttt{bfloat16} scales. All models use channel-scaled \texttt{INT8} activations.

For scalar lookup-table (LUT) formats, a single set of $K$ signed codepoints is used per weight tensor, while scales are applied per block or per output channel as shown in \cref{tab:app_quant_formats}. The \texttt{lloyd-max} baseline fits these codepoints with one-dimensional k-means/Lloyd-Max on the rescaled weight distribution. The \texttt{student-t} baseline follows \textcite{orr2025optimal}: for each tensor, we fit a Student-$t$ degrees-of-freedom parameter to the weights and choose codepoints that minimize expected squared reconstruction error under the fitted distribution.

During QAT, scalar formats are quantized in the forward pass by rescaling each block or channel and assigning each value to its nearest codepoint.

\begin{table*}[t]
\centering
\small
\caption{Scalar quantization baselines used in the format sweep. The block shape $(1,g)$ applies one scale to groups along the input dimension, while $(1,*)$ denotes per-output-channel scaling.}
\label{tab:app_quant_formats}
\begin{tabular}{@{}p{0.11\textwidth}p{0.30\textwidth}p{0.10\textwidth}p{0.11\textwidth}@{}}
\toprule
Format & Element quantizer & Block shape & $g$ \\
\midrule
\texttt{INT}
& Uniform signed integer grid
& $(1,g)$
& $\{32,64,128\}$ \\

\texttt{student-t}
& Scalar per-tensor LUT from a Student-$t$ cube-root-density quantizer
& $(1,g)$
& $\{32,64,128\}$ \\

\texttt{lloyd-max}
& Scalar per-tensor LUT fit by Lloyd-Max/k-means
& $(1,*)$
& -- \\
\bottomrule
\end{tabular}
\end{table*}

\subsection{GPTQ}
\label{sec:gptq}

\begin{figure}[t]
  \centering
  \includegraphics[width=0.8\linewidth]{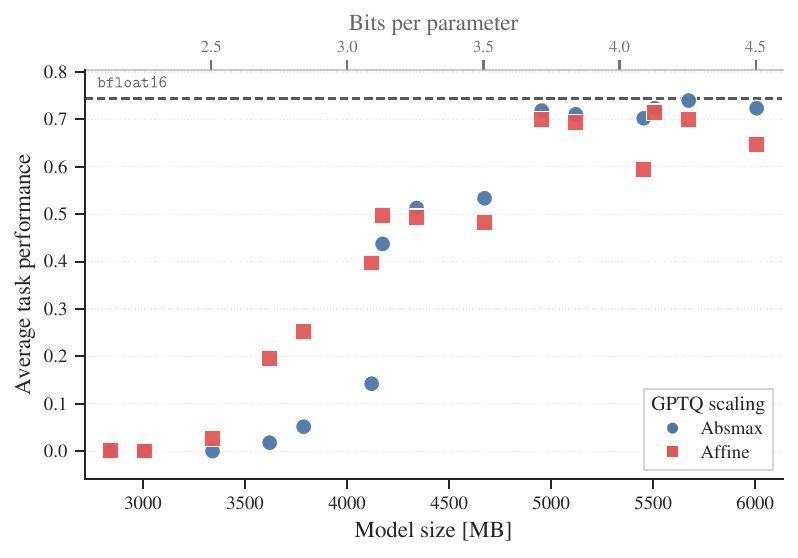}
  \caption{GPTQ scaling ablation. The absmax and affine parameterizations are evaluated using $K \in \{4,6,8,12,16\}$ and group sizes $g \in \{32,64,128\}$. The dashed line shows the original \texttt{bfloat16} model performance.}
  \label{fig:gptq-scale-ablation}
\end{figure}

We use GPTQ as a weight-only post-training quantization baseline. We implement a custom version of GPTQ that quantizes all linear weight matrices in Llama 3.2 11B Vision Instruct, including those in the vision encoder, cross-attention layers, multimodal projector, and language model output projection, whereas existing open-source libraries quantize only the text decoder using text-only calibration. All GPTQ results in this paper use our implementation.

We calibrate GPTQ using $512$ image-text samples from the synthetic dataset described in \cref{sec:data_generation}, with a maximum sequence length of $1024$ and batch size $1$. Hessian accumulation and GPTQ weight updates use \texttt{float32}. The quantized weights are saved as dense \texttt{bfloat16} values, and all activations remain in \texttt{bfloat16} during evaluation. These checkpoints therefore measure quantization accuracy rather than packed low-bit inference performance.

\paragraph{INT parameterizations}

We evaluate GPTQ with two uniform scalar \texttt{INT} parameterizations. The first is identical to the \texttt{INT} format described in \cref{sec:quant-formats}: a uniform signed integer grid containing zero, with absmax scaling and one \texttt{bfloat16} scale per group. The second follows the symmetric affine parameterization commonly used with GPTQ. It represents each quantized weight using an unsigned integer code $q \in \{0, \ldots, K-1\}$, and reconstructs the weight as $\widehat{w}=s(q-z)$, where $s$ is a \texttt{bfloat16} scale. Symmetric fitting fixes the zero point to $z=\lfloor K/2\rfloor$ for every group. Since $z$ is determined entirely by $K$, it does not require independent storage.

For the even values of $K$ used in our sweep, both parameterizations reconstruct the centered integer levels $\{-K/2,\ldots,K/2-1\}$ and differ only in their scale normalization. For a group with absmax $a$, absmax scaling uses
\[
    s_{\mathrm{absmax}}=\frac{a}{K/2-1}.
\]

Affine scaling divides the symmetric range $[-a,a]$ into $K-1$ intervals, giving
\[
    s_{\mathrm{affine}}=\frac{2a}{K-1}.
\]

For $K=16$, these scales are $a/7$ and $a/7.5$, respectively. The affine grid has finer resolution near zero, while the absmax grid represents both $\pm a$ exactly.

\Cref{fig:gptq-scale-ablation} compares the absmax and affine parameterizations: affine scaling achieves better accuracy at lower rates, while absmax scaling matches or slightly outperforms it at higher rates.

\paragraph{GPTQ with \ourformat{}}

For each linear weight matrix, we first fit \ourformat{} as described in \cref{sec:format}, using one \texttt{bfloat16} scale per output channel and one $32 \times 3$ table of integer centroids. These scales and centroids remain fixed during GPTQ. As GPTQ processes the input columns sequentially, each group of three adjacent output-channel weights within a column is normalized by the corresponding per-channel scales. The absolute values of the rescaled weights form a $3$-vector, which is assigned to the closest centroid. The resulting quantization error is then propagated to the remaining input columns using the inverse Hessian update.

\subsection{Evaluation}

Our downstream evaluation was performed on our implementations of VQAv2, DocVQA, ChartQA, and AI2D, using the VQAv2 and DocVQA validation splits, and ChartQA and AI2D test splits. Each evaluation was conducted on a fixed sample size of $1024$. For prompt templates and maximum generated sequence lengths, we followed the model guidelines \cite{llama_vision_model_card}. For implementation details we followed the guidelines in the original papers \cite{vqav2, chartqa, docvqa, ai2d}. VQAv2, ChartQA, and AI2D are evaluated on accuracy, while DocVQA uses Average Normalized Levenshtein Similarity (ANLS) with a $0.5$ threshold, as used in \textcite{mathew2022infographicvqa}.

Our absolute \texttt{bfloat16} scores are not intended to exactly reproduce the Llama 3.2 11B Vision Instruct model card numbers, as we use validation splits for VQAv2 and DocVQA, fixed $1024$-example subsets for all tasks, and our own implementation of task prompts and answer extraction and normalization. We use this evaluation as a controlled relative benchmark for quantization, with all models evaluated using the same procedure.

\subsection{Per-task prompt templates}

Following Meta's guidelines \cite{llama_vision_model_card}, we use the following prompt templates for each task.

\Needspace{8\baselineskip}\par\nobreak
\noindent\textbf{VQAv2}
\begin{lstlisting}[style=promptstyle]
Look at the image carefully and answer this visual question. For yes/no questions, just respond Yes or No. If the answer is numeric, just respond with the number and nothing else. If the answer has multiple words, just respond with the words and absolutely nothing else. Never respond in a sentence or a phrase.\n Respond with as few words as possible.\n Question: {question}
\end{lstlisting}
\Needspace{7\baselineskip}\par\nobreak
\noindent\textbf{ChartQA}
\begin{lstlisting}[style=promptstyle]
You are provided a chart image and will be asked a question. You have to think through your answer and provide a step-by-step solution. Once you have the solution, write the final answer in at most a few words at the end with the phrase "FINAL ANSWER:". The question is: {question}<cot_start>Let's think step by step.
\end{lstlisting}
\Needspace{8\baselineskip}\par\nobreak
\noindent\textbf{DocVQA}
\begin{lstlisting}[style=promptstyle]
Read the text in the image carefully and answer the question with the text as seen exactly in the image. For yes/no questions, just respond Yes or No. If the answer is numeric, just respond with the number and nothing else. If the answer has multiple words, just respond with the words and absolutely nothing else. Never respond in a sentence or a phrase.\n Question: {question}
\end{lstlisting}
\Needspace{6\baselineskip}\par\nobreak
\noindent\textbf{AI2D}
\begin{lstlisting}[style=promptstyle]
Look at the scientific diagram carefully and answer the following question: {question}\n Think step by step and finally respond to the question with only the correct option number as "FINAL ANSWER". Options:\n{options}Let's think step by step.
\end{lstlisting}

As the original template for AI2D was missing the multiple-choice options, we added the \verb|Options:\n{options}|.

\section{Additional Results}
\label{app:additional_results}

\begin{figure}[t]
  \centering
  \includegraphics[width=\columnwidth]{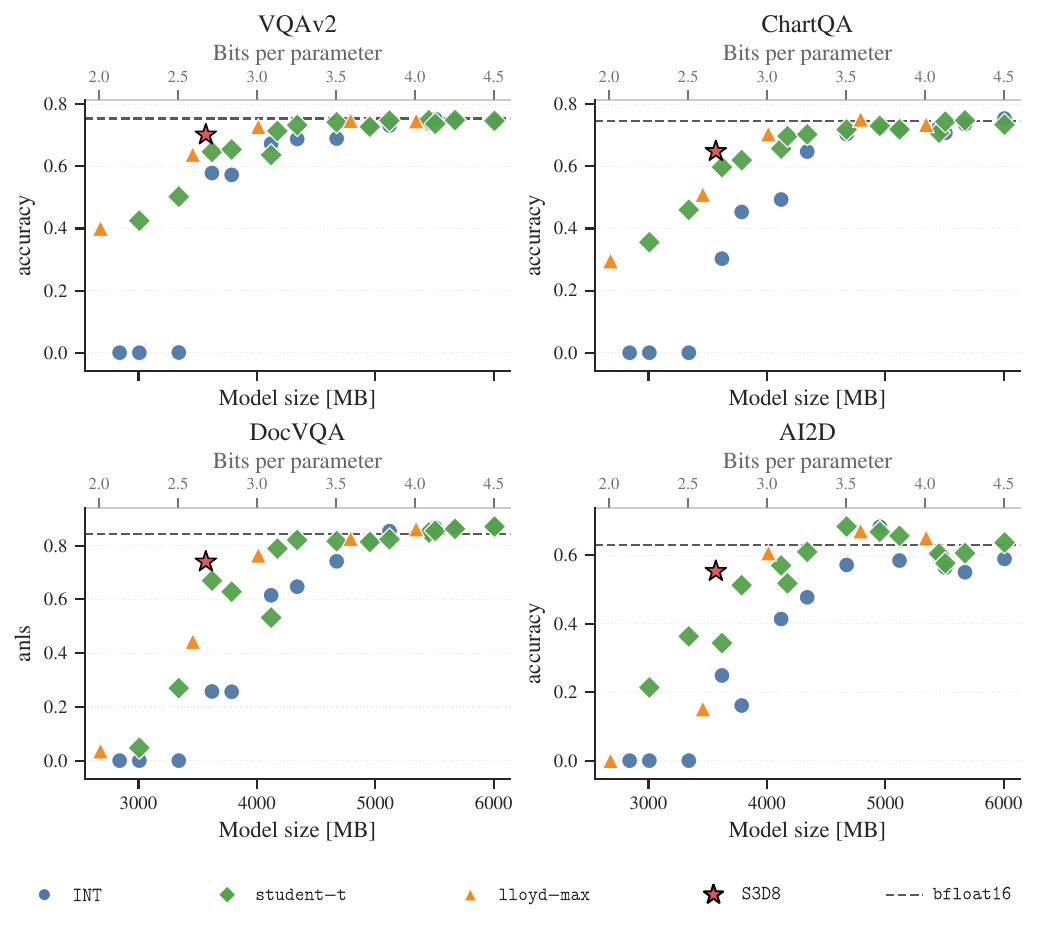}
  \caption{Individual task performance vs. model compression (QAT). The overall trend is similar to that in \cref{fig:overview}, showing a good performance-compression trade-off for \ourformat{} compared with the \texttt{INT}, \texttt{lloyd-max}, and \texttt{student-t} formats.}
  \label{fig:format_tasks}
\end{figure}

\paragraph{Task Performance}
In addition to the results in \cref{fig:overview} and \cref{fig:format_overall}, \cref{fig:format_tasks} presents the per-task trade-off between performance and model compression after QAT for different numerical formats, indicating that all tasks follow a similar trend.

\paragraph{Training Loss vs. Task Performance}

\begin{figure}[t]
  \centering
  \includegraphics[width=0.8\columnwidth]{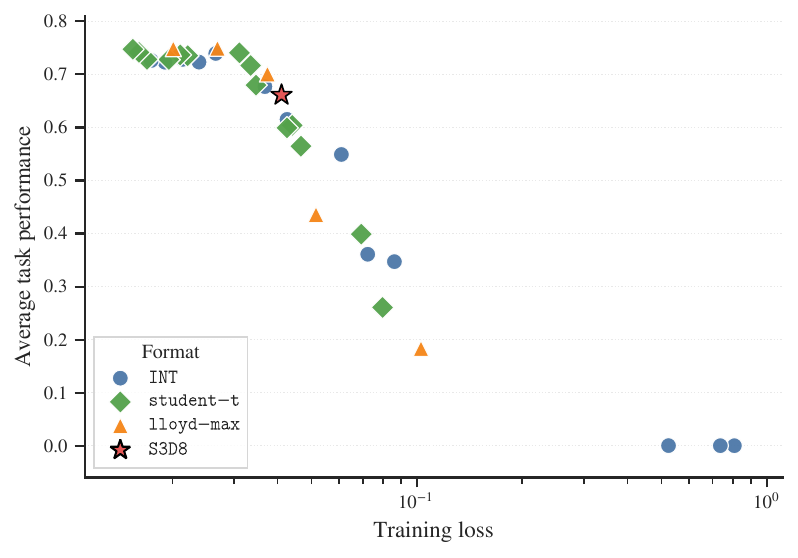}
  \caption{Training KL divergence loss versus average task performance. The figure shows a strong correlation between the training objective and final downstream performance.}
  \label{fig:task_vs_loss}
\end{figure}

\Cref{fig:task_vs_loss} shows the correlation between the KL divergence loss on the training set and the average task performance after $2048$ QAT steps. The plot shows that there is overall a strong correlation between the training objective and the final task performance (Spearman's rank correlation $\rho=-0.91$), even though the training set was chosen independently of the task data. Note that the relationship saturates at both low and high KL values: in these regimes, downstream performance changes little across models, while the loss still captures differences between them.

\paragraph{Keeping layers in \texttt{INT8}}

As an additional ablation, we evaluated \ourformat{} variants in which selected layers are kept in channel-scaled \texttt{INT8} after QAT. We hypothesized that keeping some layers in higher precision could improve downstream performance at the expense of increased model size. We tested keeping the language model output projection (\texttt{OP}), as well as the vision encoder (\texttt{VE}) in higher precision. The results are shown in \cref{tab:int8_ablation}. Keeping these layers in \texttt{INT8} reduces the training KL loss, but does not meaningfully improve downstream performance: keeping both \texttt{OP} and \texttt{VE} increases model size from $3{,}569$ MB to $4{,}513$ MB, while improving average task score by only $0.006$.

\begin{table}[t]
\centering
\caption{QAT ablation with the language-model output projection (\texttt{OP}) and vision encoder (\texttt{VE}) kept in channel-scaled \texttt{INT8}.}
\label{tab:int8_ablation}
\begin{tabular}{lrrrr}
\toprule
\texttt{INT8} layers & Size [MB] & bpp & Avg. & Train loss \\
\midrule
\texttt{None} & $3{,}569$ & $2.68$ & $0.661$ & $0.041$ \\
\texttt{OP} & $3{,}919$ & $2.94$ & $0.661$ & $0.040$ \\
\texttt{VE} & $4{,}163$ & $3.12$ & $0.666$ & $0.035$ \\
\texttt{OP + VE} & $4{,}513$ & $3.38$ & $0.667$ & $0.034$ \\
\bottomrule
\end{tabular}
\end{table}

\section{\ourformat{} Details}
\label{sec:app-format-details}

In this section, we provide format design details for \ourformat{}, further to the description of \cref{sec:format}.

\subsection{Bit packing}
\label{sec:app-bit-packing}
A straightforward bit packing for the \ourformat{} format is $\mathtt{[S_2,S_1,S_0,Q_{4..0}]}$, where the lower $5$ bits encode the centroid index and the upper $3$ bits encode the signs. However, this requires additional bit manipulation in order to extract and apply the signs after looking up the centroid.

To improve upon this, we observe that in the Arm Advanced SIMD instruction set \cite{armarch}, centroid lookup is readily implemented using \texttt{TBL}, a vectorized table lookup instruction supporting $6$-bit indices $\rightarrow$ $8$-bit values. As such, the lookup into $\mathbf{C}$ is implemented as three separate lookups into $\mathbf{C}_{\cdot,0}$, $\mathbf{C}_{\cdot,1}$, and $\mathbf{C}_{\cdot,2}$. It is therefore possible to construct \textit{signed} lookup tables, fusing the sign indicator bit into the table index to look up the signed value directly and avoid further processing.

\begin{figure}
    \centering
    \includegraphics[width=\columnwidth]{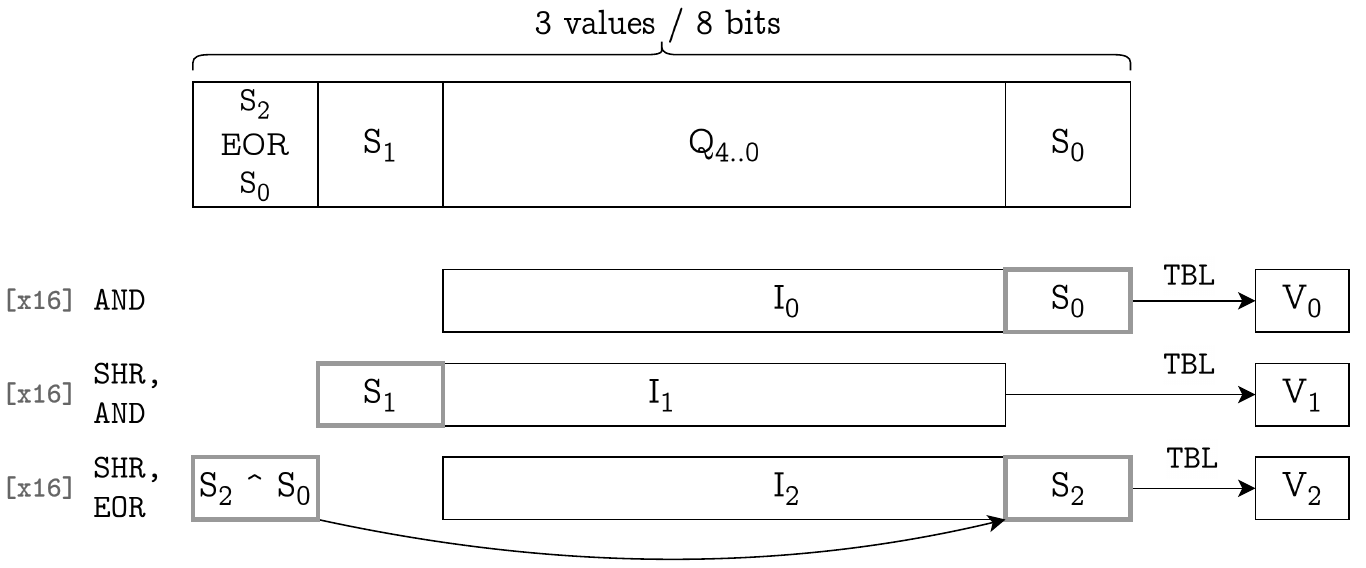}
    \caption{Bit-packed \ourformat{} storage and decoding procedure, requiring $5$ SIMD logical instructions to create indices for $3$ table lookups, decoding $48$ elements (see \cref{lst:dequant_mv_cpp}). Channel scales are typically applied later, e.g. after accumulation in a fused dequantization dot product kernel.}
    \label{fig:s3d8-bit-layout}
\end{figure}

We minimize the operations needed to construct lookup indices by packing bits as $\mathtt{[S_2\,EOR\,S_0,S_1,Q_{4..0},S_0]}$ (see \cref{fig:s3d8-bit-layout,lst:dequant_mv_cpp}).
The first index is computed by a single \texttt{AND} mask for $\mathtt{I_0=[Q_{4..0},S_0]}$.
The second index is computed by \texttt{SHR} then \texttt{AND} for $\mathtt{I_1=[S_1,Q_{4..0}]}$.
The third index is computed by \texttt{SHR} and $\texttt{EOR}\, \mathtt{I_0}$ for $\mathtt{I_2=[Q_{4..0},S_2]}$. This relies on the fact that the lookup tables may be permuted: the lookup tables for $I_0$ and $I_2$ are stored in \textit{sign-minor} order, while the lookup table for $I_1$ is stored in \textit{sign-major} order.
All operations are fully vectorizable over $16\times8$-bit elements.

Importantly, weights packed into each byte correspond to $3$ \textit{output} channels, rather than input channels. This removes the need for complicated alignment and padding along the reduction axis of a fused dequantization-dot-product kernel.

\subsection{Format design notes}
In this section, we review the design decisions behind the \ourformat{} format described in \cref{sec:format}.

\textbf{Rate ($2.7$ bits per parameter)}\quad
When quantizing weights of pretrained models using QAT, the range of $2$--$4$ bits per parameter is considered an attractive trade-off, offering small degradation from the pretrained model in exchange for high compression \cite{dettmers2023scaling,liu2025paretoq}. We can see this from our results in \cref{fig:overview}. Since our format uses short VQ, which can provide a better rate--distortion trade-off than scalar quantization, we can choose a somewhat aggressive rate of $2.7$ bits per parameter, which provides an opportunity for advanced QAT techniques to improve accuracy.

\textbf{Compute format (\texttt{INT8})}\quad
The \texttt{INT8} format enjoys compute support in the Arm \texttt{dotprod} and \texttt{i8mm} extensions, which are broadly supported in modern mobile CPUs. As the fastest format with wide availability at the time of this paper, it generally provides substantially higher arithmetic throughput than \texttt{bfloat16} or \texttt{float16}.

\textbf{Vector Quantization}\quad
The Arm SIMD instruction set includes \texttt{TBL}, which selects single bytes from a lookup table of $16$, $32$, $48$ or $64$ elements. If using scalar quantization, a common optimization is to form larger vector lookup tables rather than separately masking and shifting small indices. These vector lookup tables are the Cartesian product of the scalar centroids, so that in this case there is no runtime cost to replacing these with vector centroids.

\textbf{Absolute-Value Centroids \& Sign Indicator Bits}\quad
The most straightforward way to encode $3$ values in an $8$-bit vector format would be to encode a single $8$-bit centroid index into a lookup table of shape $(256, 3)$. However, this $768$-byte lookup table is too large for SIMD registers. Since we expect the main benefit of VQ to come from shaping centroids to the distribution, not from particular covariance structures, we can exploit the symmetry expected and specify our $256$ centroids as all $8$ reflections of $32$ absolute-value centroids. This allows for a lookup table of shape $(32, 3)$ with separate sign indicator bits (but note that we revise this lookup table layout with our next design decision). NSNQuant \cite{son2025nsnquant} uses a related sign-factorized construction for $8$-dimensional KV-cache vector quantization, where separate sign bits accompany the codebook index.

\textbf{Fused sign lookups}\quad
A straightforward implementation of the format would look up the absolute value of each vector element, then unpack and apply the sign indicator bit for that element. We are not aware of a fast way of doing this in Arm Advanced SIMD (simple approaches would use $4$ instructions for each of the $3$ elements to shift and apply two's complement). Our approach trades register pressure for instruction count by constructing lookup tables containing signed values, requiring $3\times64$-byte lookup tables in total.

\textbf{Bit layout}\quad
We save $1$ to $2$ instructions by adopting the bit layout described in \cref{sec:app-bit-packing}, using permuted lookup tables. Since weight storage bit-packing and lookup table permutation can be performed once before loading the model, there is no runtime cost to doing this.

\textbf{Output channel packing}\quad
While it might seem natural to pack elements of $W \in \mathbb{R}^{n \times k}$ that are consecutive on the reduction $k$-axis into a single byte, this would require extra instructions to interleave weight elements, or unzip them from the input $x \in \mathbb{R}^{k}$. Since kernel implementations that exploit instruction-level parallelism would typically compute multiple outputs in a single pass through the inputs, it is straightforward to pack elements corresponding to consecutive output channels in each byte. This means that the only padding required is $3\cdot\lceil n/3 \rceil - n$ on the output channels.

\section{Runtime Performance Benchmarking}
\label{sec:app-benchmarking}

We evaluate C++ implementations of blocked matrix products and fused matrix products, with extensive use of template code generation to exploit instruction-level parallelism, and using OpenMP \cite{openmp} for outer-loop parallelism.

Each benchmark sample executes $20$ iterations with distinct copies of the inputs to ensure that operands come from memory rather than on-chip cache. We discard the first $5$ iteration times and record the mean wall-clock time of the remaining $15$ iterations as the sample. We cycle through benchmarks, collecting $50$ such samples from each. We observe asymmetric (slow) outliers for these samples, which we attribute to non-deterministic OS scheduling, CPU frequency scaling and memory management. We therefore report the median of the samples and use bootstrapping \cite{efron1979bootstrap} to calculate a 95\% confidence interval.

\begin{figure}[!t]
    \centering
    \includegraphics[width=\columnwidth]{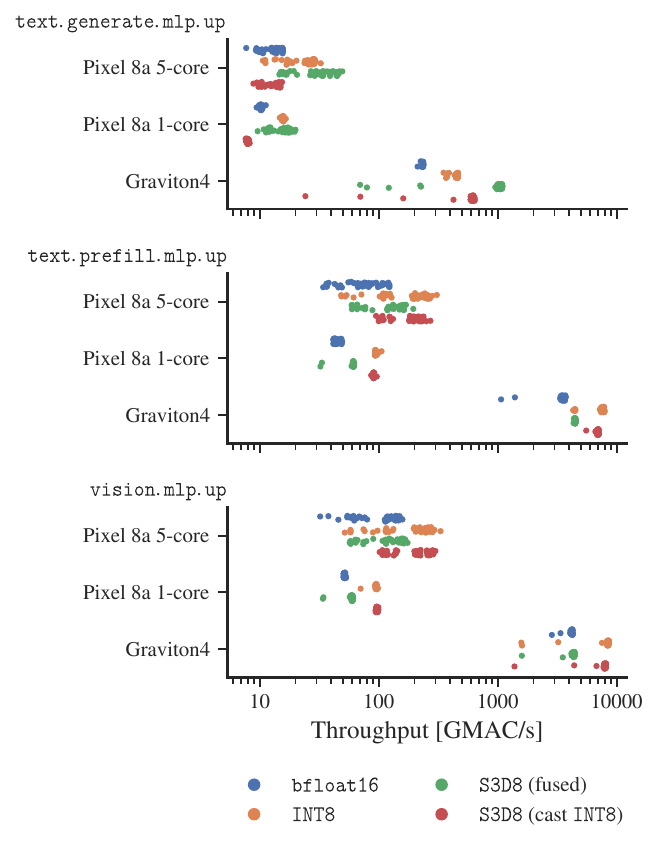}
    \caption{
    Raw matrix multiply benchmark sample measurements for $3$ selected cases, across devices and formats. We note that Graviton4 and Pixel 8a $1$-core throughput measurements are generally very stable, with some low-throughput outliers, while Pixel 8a $5$-core measurements have high variance, perhaps due to scheduling or thermals. In the generation case, fused \ourformat{} is faster than \texttt{INT8} on Graviton4 and the five-core Pixel, with similar throughput on the single-core Pixel. For the prefill cases, unfused ``cast'' \ourformat{} is modestly slower than \texttt{INT8} on Graviton4, with similar throughput on the five-core Pixel. See \cref{tbl:bench-matmul-android-mt,tbl:bench-matmul-android-1t,tbl:bench-matmul-graviton} for aggregate measurements on the full set of cases.
    }
    \label{fig:bench-matmul-samples}
\end{figure}

\paragraph{Pixel 8a} As a target mobile platform, we selected the Google Android Pixel 8a (2024), which has $1$ ``prime'' Cortex-X3 core, $4$ ``performance'' Cortex-A715 cores and $4$ ``efficiency'' Cortex-A510 cores. We observed very high variance in timing measurements when using $9$ or $5$ cores with OpenMP, although this had the potential for the best performance. We therefore evaluate two configurations: \textit{Pixel 8a $5$-core} using \texttt{OMP\_NUM\_THREADS=5} and \texttt{taskset 1f0} (prime + performance), and \textit{Pixel 8a $1$-core} using \texttt{taskset 100} and \texttt{OMP\_NUM\_THREADS=1} to restrict execution to the prime Cortex-X3 core, which provided more stable timings (see \cref{fig:bench-matmul-samples}). The end-to-end $3.8$ tokens/s result uses the $5$-core configuration: $5$ OpenMP threads restricted to the prime and performance cores. For both kernel benchmarks and end-to-end measurements, we removed the case, placed the phone on a cool thermally conductive surface, disabled adaptive battery and enabled airplane mode.

Details:

{
\small
\renewcommand{\arraystretch}{1.3}
\vspace{0.5em}
\begin{tabular}{lp{5cm}}
\toprule
Parameter & Value \\
\midrule
Device             & Pixel 8a \\
CPU                & $1\times$ Cortex-X3 + $4\times$ Cortex-A715 + $4\times$ Cortex-A510 \\
Memory             & $4\times$ LPDDR5X-4266 ($34$ GB/s) \\
OS                 & Android 14 (Linux 6.1.145) \\
Compiler           & clang 21.0.0 (NDK 30.0.14904198) \\
Target             & \texttt{aarch64-linux-android35} \\
Flags              & \texttt{-O3 -march=armv9-a+bf16+i8mm -std=c++20 -DNDEBUG -fopenmp} \\
\hline
\multicolumn{2}{l}{\textit{Pixel 8a $5$-core}} \\
\texttt{OMP\_NUM\_THREADS} & $5$ (\texttt{taskset 1f0}) \\
\hline
\multicolumn{2}{l}{\textit{Pixel 8a $1$-core}} \\
\texttt{OMP\_NUM\_THREADS} & $1$ (\texttt{taskset 100}) \\
\bottomrule
\end{tabular}
}

\paragraph{Graviton4} Our server CPU benchmarking details are as follows:

{
\small
\renewcommand{\arraystretch}{1.3}
\vspace{0.5em}
\begin{tabular}{lp{5cm}}
\toprule
Parameter & Value \\
\midrule
AWS Instance       & c8g.24xlarge \\
CPU                & Graviton4 ($96\times$ Neoverse V2), $36$ MiB LLC \\
Memory             & $12\times$ DDR5-5600 ($538$ GB/s) \\
OS                 & Ubuntu 24.04.3 \\
OpenMP version     & 5.1 \\
Compiler           & clang 20.1.2 \\
Flags              & \texttt{-O3 -mcpu=native -std=c++20 -DNDEBUG -fopenmp} \\
\texttt{OMP\_NUM\_THREADS} & $96$ \\
\bottomrule
\end{tabular}
}

\newpage
\makeatletter
\setlength{\@fptop}{0pt}
\makeatother

\begin{figure}[h]
    \centering
    \includegraphics[width=\columnwidth]{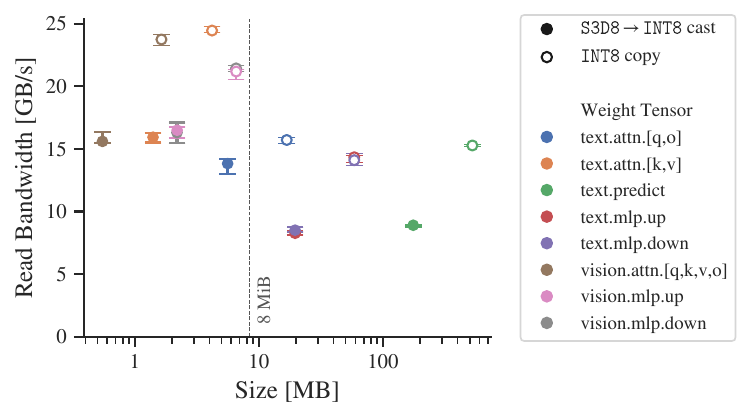}
    \caption{$\ourformat{} \rightarrow \texttt{INT8}$ dequantization kernel performance on the Pixel 8a, using $5$ cores. The benchmark cycles through multiple input tensors to avoid read cache hits, but always writes to the same destination to encourage write hits. We report \textit{read bandwidth}, with respect to the source tensor size only, as destination writes may remain in the cache (only possible in practice for outputs smaller than the L2 cache size marked). For \ourformat{} dequantization, note that write bandwidth is $3\times$ read bandwidth, while for the \texttt{INT8} copy baseline, write bandwidth is the same as read bandwidth.
    The kernel is able to achieve an appreciable fraction of the bandwidth achieved by the baseline, due to the simplicity of \ourformat{} decoding. Note, therefore, that the wall clock time of an \ourformat{} cast is lower than an \texttt{INT8} copy for a given weight tensor, since it reads $\frac{1}{3}$ as many bytes.
    Error bars show the $95\%$ bootstrapping confidence interval for the median.
    }
    \label{fig:bench-android-dequant}
\end{figure}

\begin{figure}[h]
    \centering
    \includegraphics[width=\columnwidth]{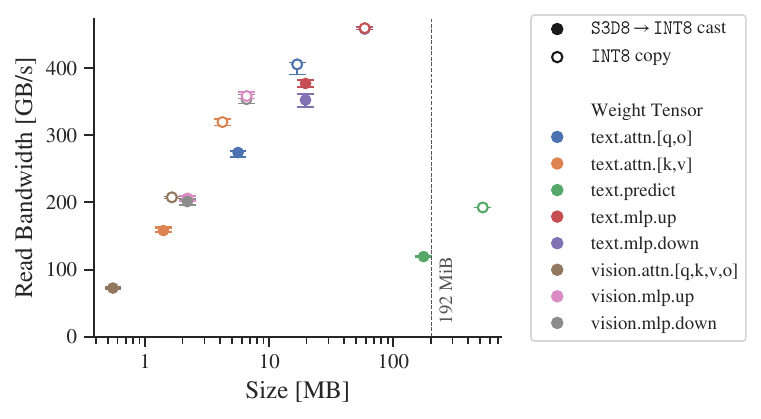}
    \caption{$\ourformat{} \rightarrow \texttt{INT8}$ dequantization kernel performance on a $96$-core Graviton4 CPU. The setup follows that of \cref{fig:bench-android-dequant}. Read bandwidth increases as the output tensor size approaches the total L2 cache size, after which the need to write the output to memory reduces the achievable read bandwidth.
    Error bars show the $95\%$ bootstrapping confidence interval for the median.
    }
    \label{fig:bench-graviton-dequant}
\end{figure}

\begin{table}[h]
    \caption{Compute rate of selected matrix multiply shapes on the Pixel 8a $5$-core.
    The first group of operations with $m\!=\!1$ corresponds to autoregressive text generation, for \texttt{attn.[q,o]}, \texttt{attn.[k,v]}, \texttt{mlp.up}, \texttt{mlp.down} and \texttt{predict}. The second group with $m\!=\!128$ corresponds to the same layers, but for a short prefill of $128$ tokens (\texttt{predict} is excluded as this is not needed during prefill). The final group with $m\!=\!1601$ corresponds to vision prefill of a single tile ($40\times40+1$ tokens), for \texttt{attn.[q,k,v,o]}, \texttt{mlp.up} and \texttt{mlp.down} respectively. Error ranges show the 95\% bootstrapping confidence interval for the median. See \cref{fig:bench-matmul-samples} for an example of the sample distribution.
    }
    \label{tbl:bench-matmul-android-mt}
    \centering
    \small
\begin{tabular}{p{2.5cm} C{1.4cm} C{1.4cm} C{1.4cm}}
\toprule
Shape & \multicolumn{3}{c}{GMAC/s} \\
$(m,k,n)$ & $\mathtt{bfloat16}$ & $\mathtt{INT8}$ & $\mathtt{S3D8}$ \\
\midrule
$(1, 4096, 4096)$ & \errorrange{$14.2$}{$[12.9,14.9]$} & \errorrange{$\mathbf{29.2}$}{$[26.6,30.6]$} & \errorrange{$\mathbf{31.2}$}{$[24.7,34.4]$} \\
$(1, 4096, 1024)$ & \errorrange{$12.2$}{$[12.0,12.8]$} & \errorrange{$\mathbf{29.2}$}{$[25.5,29.9]$} & \errorrange{$\mathbf{30.2}$}{$[28.0,32.6]$} \\
$(1, 4096, 14336)$ & \errorrange{$13.6$}{$[13.1,13.9]$} & \errorrange{$26.5$}{$[22.0,27.9]$} & \errorrange{$\mathbf{33.8}$}{$[29.8,36.7]$} \\
$(1, 14336, 4096)$ & \errorrange{$13.2$}{$[12.3,13.3]$} & \errorrange{$23.7$}{$[17.9,24.5]$} & \errorrange{$\mathbf{34.3}$}{$[31.2,35.1]$} \\
$(1, 4096, 128256)$ & \errorrange{$11.8$}{$[11.5,13.0]$} & \errorrange{$\mathbf{28.1}$}{$[25.7,31.3]$} & \errorrange{$\mathbf{31.7}$}{$[23.4,37.7]$} \\
\midrule
$(128, 4096, 4096)$ & \errorrange{$88.3$}{$[78.5,97.3]$} & \errorrange{$\mathbf{205.5}$}{$[128.8,218.6]$} & \errorrange{$\mathbf{202.1}$}{$[190.0,212.6]$} \\
$(128, 4096, 1024)$ & \errorrange{$79.9$}{$[72.4,95.1]$} & \errorrange{$\mathbf{204.6}$}{$[136.0,218.7]$} & \errorrange{$\mathbf{213.8}$}{$[201.5,229.4]$} \\
$(128, 4096, 14336)$ & \errorrange{$76.3$}{$[72.1,83.5]$} & \errorrange{$\mathbf{201.1}$}{$[190.4,214.4]$} & \errorrange{$\mathbf{197.7}$}{$[183.2,207.7]$} \\
$(128, 14336, 4096)$ & \errorrange{$107.8$}{$[105.5,115.9]$} & \errorrange{$\mathbf{214.3}$}{$[124.2,227.5]$} & \errorrange{$\mathbf{203.6}$}{$[185.5,215.8]$} \\
\midrule
$(1601, 1280, 1280)$ & \errorrange{$122.9$}{$[116.3,127.9]$} & \errorrange{$\mathbf{212.3}$}{$[139.5,226.5]$} & \errorrange{$\mathbf{229.9}$}{$[208.8,239.1]$} \\
$(1601, 1280, 5120)$ & \errorrange{$116.5$}{$[80.0,121.3]$} & \errorrange{$\mathbf{206.1}$}{$[159.9,234.0]$} & \errorrange{$\mathbf{211.8}$}{$[203.8,222.9]$} \\
$(1601, 5120, 1280)$ & \errorrange{$118.6$}{$[115.0,126.4]$} & \errorrange{$\mathbf{230.0}$}{$[218.8,240.7]$} & \errorrange{$\mathbf{238.6}$}{$[219.3,264.5]$} \\
\bottomrule
\end{tabular}
\end{table}

\begin{table}[h]
    \caption{Compute rate of selected matrix multiply shapes on the Pixel 8a $1$-core. \textit{See \cref{tbl:bench-matmul-android-mt} for further details.}
    }
    \label{tbl:bench-matmul-android-1t}
    \centering
    \small
\begin{tabular}{p{2.5cm} C{1.4cm} C{1.4cm} C{1.4cm}}
\toprule
Shape & \multicolumn{3}{c}{GMAC/s} \\
$(m,k,n)$ & $\mathtt{bfloat16}$ & $\mathtt{INT8}$ & $\mathtt{S3D8}$ \\
\midrule
$(1, 4096, 4096)$ & \errorrange{$10.4$}{$[10.2,10.6]$} & \errorrange{$15.3$}{$[15.2,15.3]$} & \errorrange{$\mathbf{19.3}$}{$[18.8,19.9]$} \\
$(1, 4096, 1024)$ & \errorrange{$11.2$}{$[11.1,11.2]$} & \errorrange{$14.7$}{$[14.5,14.9]$} & \errorrange{$\mathbf{19.9}$}{$[19.4,20.4]$} \\
$(1, 4096, 14336)$ & \errorrange{$10.3$}{$[10.2,10.3]$} & \errorrange{$\mathbf{15.7}$}{$[15.6,15.7]$} & \errorrange{$\mathbf{16.0}$}{$[14.9,17.0]$} \\
$(1, 14336, 4096)$ & \errorrange{$11.1$}{$[11.1,11.1]$} & \errorrange{$\mathbf{22.2}$}{$[22.1,22.3]$} & \errorrange{$16.5$}{$[13.4,17.5]$} \\
$(1, 4096, 128256)$ & \errorrange{$10.3$}{$[10.3,10.4]$} & \errorrange{$15.8$}{$[15.8,15.8]$} & \errorrange{$\mathbf{16.4}$}{$[16.3,16.6]$} \\
\midrule
$(128, 4096, 4096)$ & \errorrange{$44.7$}{$[43.0,47.4]$} & \errorrange{$\mathbf{96.0}$}{$[95.8,96.3]$} & \errorrange{$93.7$}{$[93.5,94.0]$} \\
$(128, 4096, 1024)$ & \errorrange{$46.1$}{$[44.3,48.0]$} & \errorrange{$\mathbf{94.4}$}{$[91.5,95.6]$} & \errorrange{$\mathbf{95.4}$}{$[93.9,96.7]$} \\
$(128, 4096, 14336)$ & \errorrange{$43.2$}{$[42.7,44.6]$} & \errorrange{$\mathbf{93.8}$}{$[93.7,94.1]$} & \errorrange{$90.6$}{$[90.4,90.7]$} \\
$(128, 14336, 4096)$ & \errorrange{$48.8$}{$[48.6,48.9]$} & \errorrange{$\mathbf{97.3}$}{$[97.1,97.7]$} & \errorrange{$92.0$}{$[91.6,92.4]$} \\
\midrule
$(1601, 1280, 1280)$ & \errorrange{$51.5$}{$[51.3,51.8]$} & \errorrange{$96.9$}{$[96.6,97.1]$} & \errorrange{$\mathbf{97.9}$}{$[97.7,98.1]$} \\
$(1601, 1280, 5120)$ & \errorrange{$51.4$}{$[51.4,51.7]$} & \errorrange{$95.5$}{$[95.2,95.7]$} & \errorrange{$\mathbf{96.9}$}{$[96.6,97.1]$} \\
$(1601, 5120, 1280)$ & \errorrange{$50.6$}{$[50.5,50.7]$} & \errorrange{$101.2$}{$[101.1,101.4]$} & \errorrange{$\mathbf{102.4}$}{$[102.2,102.6]$} \\
\bottomrule
\end{tabular}
\end{table}

\begin{table}[!t]
    \caption{Compute rate of selected matrix multiply shapes on a $96$-core Graviton4 CPU. \textit{See \cref{tbl:bench-matmul-android-mt} for further details.}
    }
    \label{tbl:bench-matmul-graviton}
    \centering
    \small
\begin{tabular}{p{2.5cm} C{1.4cm} C{1.4cm} C{1.4cm}}
\toprule
Shape & \multicolumn{3}{c}{GMAC/s} \\
$(m,k,n)$ & $\mathtt{bfloat16}$ & $\mathtt{INT8}$ & $\mathtt{S3D8}$ \\
\midrule
$(1, 4096, 4096)$ & \errorrange{$195$}{$[194,196]$} & \errorrange{$406$}{$[402,408]$} & \errorrange{$\mathbf{694}$}{$[682,705]$} \\
$(1, 4096, 1024)$ & \errorrange{$157$}{$[155,157]$} & \errorrange{$294$}{$[290,298]$} & \errorrange{$\mathbf{386}$}{$[373,390]$} \\
$(1, 4096, 14336)$ & \errorrange{$236$}{$[235,236]$} & \errorrange{$461$}{$[458,461]$} & \errorrange{$\mathbf{1031}$}{$[1021,1047]$} \\
$(1, 14336, 4096)$ & \errorrange{$226$}{$[225,227]$} & \errorrange{$458$}{$[456,458]$} & \errorrange{$\mathbf{965}$}{$[961,970]$} \\
$(1, 4096, 128256)$ & \errorrange{$253$}{$[252,253]$} & \errorrange{$541$}{$[540,550]$} & \errorrange{$\mathbf{1592}$}{$[1562,1636]$} \\
\midrule
$(128, 4096, 4096)$ & \errorrange{$3471$}{$[3362,3517]$} & \errorrange{$\mathbf{7278}$}{$[7155,7370]$} & \errorrange{$6068$}{$[5994,6104]$} \\
$(128, 4096, 1024)$ & \errorrange{$2889$}{$[2875,2909]$} & \errorrange{$\mathbf{5852}$}{$[5816,5881]$} & \errorrange{$4624$}{$[4572,4660]$} \\
$(128, 4096, 14336)$ & \errorrange{$3492$}{$[3466,3522]$} & \errorrange{$\mathbf{7700}$}{$[7614,7738]$} & \errorrange{$6994$}{$[6973,7008]$} \\
$(128, 14336, 4096)$ & \errorrange{$3168$}{$[3126,3180]$} & \errorrange{$\mathbf{5578}$}{$[5409,5719]$} & \errorrange{$5169$}{$[4975,5239]$} \\
\midrule
$(1601, 1280, 1280)$ & \errorrange{$4111$}{$[4085,4126]$} & \errorrange{$\mathbf{8221}$}{$[8183,8230]$} & \errorrange{$6685$}{$[6669,6760]$} \\
$(1601, 1280, 5120)$ & \errorrange{$4210$}{$[4193,4224]$} & \errorrange{$\mathbf{8464}$}{$[8422,8508]$} & \errorrange{$7979$}{$[7947,8006]$} \\
$(1601, 5120, 1280)$ & \errorrange{$4277$}{$[4267,4296]$} & \errorrange{$\mathbf{8743}$}{$[8655,8816]$} & \errorrange{$8112$}{$[8065,8155]$} \\
\bottomrule
\end{tabular}
\end{table}


\lstdefinelanguage{ARM}{
  morekeywords={ldr,add,cmp,and,ushr,eor,tbl,sdot,b.hi},
  sensitive=true,
  alsoletter={.},
  morecomment=[l]{//},
  keywordstyle=\color{purple},
  literate={}
}

\lstdefinestyle{cpp-neon}{
  language=C++,
  deletekeywords={float},
  morekeywords=[2]{
    float,__bf16,int8_t,uint8_t,uint64_t,
    int8x16_t,uint8x16_t,int32x4_t,int8x16x4_t
  },
  keywordstyle=[2]\color{purple},
  morekeywords=[3]{
    vdupq_n_s32,vld1q_s8,vld1q_u8,
    vandq_u8,vshrq_n_u8,veorq_u8,
    vqtbl4q_s8,vdotq_s32,vaddvq_s32,
    vcvth_bf16_f32,vcvtah_f32_bf16,vdupq_n_u8
  },
  keywordstyle=[3]\color{teal},
  literate={}
}


\begin{figure*}[p]
\centering
\begin{lstlisting}[
  caption={
  A simplified example implementation of a fused \ourformat{} dequantization matrix-vector product with channel scaling of both inputs \texttt{x} and weight. Decoding follows the procedure outlined in \cref{fig:s3d8-bit-layout}, accumulation is performed in \texttt{int32} with input and weight channel scales applied together before writing a \texttt{bfloat16} output. Our practical benchmark implementation is similar, but constructs a larger block of accumulators to exploit instruction-level parallelism.
  },
  label={lst:dequant_mv_cpp},
  language=C++,
  style=cpp-neon,
]
void mv_s3d8(const int8_t* x,
             const float x_s,
             const uint8_t* W_q,
             const int8x16x4_t* W_C,
             const __bf16* W_s,
             const uint64_t k,
             __bf16* y) {
    int32x4_t acc[3] = {vdupq_n_s32(0), vdupq_n_s32(0), vdupq_n_s32(0)};

    for (auto ik = 0u; ik < k; ik += 16) {
        int8x16_t xk = vld1q_s8(&x[ik]);
        uint8x16_t pack = vld1q_u8(&W_q[ik]);

        uint8x16_t I0 = vandq_u8(pack, vdupq_n_u8(0x3F));
        uint8x16_t I1 = vandq_u8(vshrq_n_u8(pack, 1), vdupq_n_u8(0x3F));
        uint8x16_t I2 = veorq_u8(I0, vshrq_n_u8(pack, 7));
        int8x16_t V0 = vqtbl4q_s8(W_C[0], I0);
        int8x16_t V1 = vqtbl4q_s8(W_C[1], I1);
        int8x16_t V2 = vqtbl4q_s8(W_C[2], I2);

        acc[0] = vdotq_s32(acc[0], xk, V0);
        acc[1] = vdotq_s32(acc[1], xk, V1);
        acc[2] = vdotq_s32(acc[2], xk, V2);
    }

    y[0] = vcvth_bf16_f32(x_s * float(vaddvq_s32(acc[0])) * vcvtah_f32_bf16(W_s[0]));
    y[1] = vcvth_bf16_f32(x_s * float(vaddvq_s32(acc[1])) * vcvtah_f32_bf16(W_s[1]));
    y[2] = vcvth_bf16_f32(x_s * float(vaddvq_s32(acc[2])) * vcvtah_f32_bf16(W_s[2]));
}
\end{lstlisting}

\begin{lstlisting}[
  caption={
  Compiler disassembly for the fused \ourformat{} dequantization matrix-vector product ``hot loop'' from \cref{lst:dequant_mv_cpp}. Note that in {\color{gray}\texttt{format decoding}}, $5$ logical instructions are used to generate indices and $3$ for \texttt{tbl} lookups, to decode $48$ elements.
  },
  label={lst:dequant_mv_s},
  language=ARM,
]
.LBB184_5:
    ldr  q26, [x2, x1]                          // Loading/looping
    ldr  q25, [x15, x1]
    add  x1, x1, #16
    cmp  x12, w1, uxtw

    and  v27.16b, v26.16b, v5.16b               // Format decoding
    ushr v28.16b, v26.16b, #1
    and  v28.16b, v28.16b, v5.16b
    ushr v26.16b, v26.16b, #7
    eor  v26.16b, v27.16b, v26.16b
    tbl  v29.16b, { v1.16b, v2.16b, v3.16b, v4.16b }, v27.16b
    tbl  v28.16b, { v16.16b, v17.16b, v18.16b, v19.16b }, v28.16b
    tbl  v26.16b, { v20.16b, v21.16b, v22.16b, v23.16b }, v26.16b

    sdot v24.4s, v25.16b, v29.16b               // Dot product
    sdot v7.4s, v25.16b, v28.16b
    sdot v6.4s, v25.16b, v26.16b

    b.hi .LBB184_5
\end{lstlisting}
\end{figure*}

\clearpage

\makeatletter
\setlength{\@dblfptop}{0pt}
\setlength{\@dblfpsep}{60pt plus 2pt minus 2pt}
\setlength{\@dblfpbot}{0pt plus 1fil}
\makeatother

\lstdefinestyle{model-output}{
  basicstyle=\ttfamily\fontsize{7}{8}\selectfont,
  columns=fullflexible,
  keepspaces=true,
  breaklines=true,
  breakatwhitespace=false,
  breakautoindent=false,
  breakindent=0pt,
  showstringspaces=false,
  aboveskip=2pt,
  belowskip=4pt
}

\begin{figure*}[t]
\section{Model Generation Examples}
\label{app:generation_examples}

\Cref{fig:qual-vqav2,fig:qual-chartqa,fig:qual-docvqa,fig:qual-ai2d} show one
example from each downstream task, comparing generations from the original
\texttt{bfloat16} model and the \ourformat{} model under the same evaluation
prompt and decoding setup.

\vspace{1.1em}
\centering
\begin{minipage}[t]{0.34\textwidth}
\vspace{0pt}
\centering
\includegraphics[width=\linewidth,height=0.34\textheight,keepaspectratio]{figs/vqa_example.png}
\end{minipage}\hfill
\begin{minipage}[t]{0.62\textwidth}
\vspace{0pt}
\raggedright
\small
\textbf{Task:} VQAv2

\smallskip
\textbf{Question:} How many computers?

\smallskip
\textbf{Reference answers:} 4, 4, 6, 4, 5, 4, 5, 5, 5, 5

\medskip
\textbf{\texttt{bfloat16}:}
\begin{lstlisting}[style=model-output]
5
\end{lstlisting}

\medskip
\textbf{\ourformat{}:}
\begin{lstlisting}[style=model-output]
4
\end{lstlisting}
\end{minipage}
\caption{VQAv2 example. The human answers contain both ``4'' and
``5'', so both generations match the evaluation target.}
\label{fig:qual-vqav2}
\end{figure*}

\begin{figure*}[t]
\centering
\begin{minipage}[t]{0.34\textwidth}
\vspace{0pt}
\centering
\includegraphics[width=\linewidth,height=0.34\textheight,keepaspectratio]{figs/chartqa_example.png}
\end{minipage}\hfill
\begin{minipage}[t]{0.62\textwidth}
\vspace{0pt}
\raggedright
\footnotesize
\textbf{Task:} ChartQA

\smallskip
\textbf{Question:} What is the average of all the bars?

\smallskip
\textbf{Reference answer:} 35

\medskip
\textbf{\texttt{bfloat16}:}
\begin{lstlisting}[style=model-output]
To calculate the average of the given bars, the following steps are taken:

* Add up the values of all the bars: 19.0, 39.0, and 47.0.
* Calculate the sum: 19.0 + 39.0 + 47.0 = 105.0
* Divide the sum by the number of bars: 105.0 / 3 = 35.0

*Answer*: 35.0
\end{lstlisting}

\medskip
\textbf{\ourformat{}:}
\begin{lstlisting}[style=model-output]
To find the average of all the bars, we need to add up the values of all the bars and then divide by the total number of bars.

The values of the bars are: 19.0, 39.0, and 47.0.

First, let's add up these values: 19.0 + 39.0 + 47.0 = 105.0

Next, we divide this sum by the total number of bars, which is 3: 105.0 / 3 = 35.0

Therefore, the average of all the bars is 35.0.

FINAL ANSWER: 35.0
\end{lstlisting}
\end{minipage}
\caption{ChartQA example. Both models recover the correct final answer.}
\label{fig:qual-chartqa}
\end{figure*}

\begin{figure*}[t]
\centering
\begin{minipage}[t]{0.34\textwidth}
\vspace{0pt}
\centering
\includegraphics[width=\linewidth,height=0.40\textheight,keepaspectratio]{figs/docvqa_example.png}
\end{minipage}\hfill
\begin{minipage}[t]{0.62\textwidth}
\vspace{0pt}
\raggedright
\small
\textbf{Task:} DocVQA

\smallskip
\textbf{Question:} What type of report is this?

\smallskip
\textbf{Reference answer:} Preliminary

\medskip
\textbf{\texttt{bfloat16}:}
\begin{lstlisting}[style=model-output]
This is a preliminary report.
\end{lstlisting}

\medskip
\textbf{\ourformat{}:}
\begin{lstlisting}[style=model-output]
Nutrition survey.
\end{lstlisting}
\end{minipage}
\caption{DocVQA example. The quantized model does not generate the reference answer, which the \texttt{bfloat16} model does.}
\label{fig:qual-docvqa}
\end{figure*}

\begin{figure*}[t]
\centering
\begin{minipage}[t]{0.34\textwidth}
\vspace{0pt}
\centering
\includegraphics[width=\linewidth,height=0.32\textheight,keepaspectratio]{figs/ai2d_example.png}
\end{minipage}\hfill
\begin{minipage}[t]{0.62\textwidth}
\vspace{0pt}
\raggedright
\footnotesize
\textbf{Task:} AI2D

\smallskip
\textbf{Question:} Which of these is used for locomotion or support?
Options: 1) Antenna, 2) Abdomen, 3) Leg, 4) Head.

\smallskip
\textbf{Reference answer:} 3) Leg

\medskip
\textbf{\texttt{bfloat16}:}
\begin{lstlisting}[style=model-output]
To determine which part is used for locomotion or support, we need to consider the functions of each option. The abdomen is primarily involved in digestion and storage of food, the head contains the brain and sensory organs, and the antenna is used for sensing the environment. However, the leg is directly involved in movement and support.

**Correct option: 3) Leg**
\end{lstlisting}

\medskip
\textbf{\ourformat{}:}
\begin{lstlisting}[style=model-output]
To determine which part of the insect is used for locomotion or support, we need to consider the functions of each option provided.

*   **Antenna**: The antenna is used for sensing the environment and detecting pheromones, not for locomotion or support.
*   **Abdomen**: The abdomen is the body part that contains the digestive system, reproductive organs, and muscles used for movement. It is not directly involved in locomotion or support.
*   **Legs**: Insects use their legs for walking, running, jumping, and grasping. This makes legs a crucial part of their locomotion system.
*   **Head**: The head contains the brain, eyes, mouthparts, and other sensory organs. While it plays a role in the overall function of the insect, it is not directly involved in locomotion or support.

Given the information above, the correct answer is:

**FINAL ANSWER: 3) Leg**
\end{lstlisting}
\end{minipage}
\caption{AI2D example. Both models select the correct multiple-choice answer, with the quantized model producing a longer rationale.}
\label{fig:qual-ai2d}
\end{figure*}

\end{document}